\pdfoutput=1
\documentclass[lettersize,journal]{IEEEtran}

\usepackage{times}
\usepackage{graphics} %
\usepackage{graphicx}
\usepackage{subfigure}
\usepackage{amsmath,amssymb,amsopn,amstext,amsfonts}
\usepackage{cancel}
\usepackage[space]{cite}
\usepackage{pdfsync}
\usepackage{balance}
\usepackage{color}
\usepackage{mathtools}
\usepackage{bm}

\usepackage{diagbox}
\usepackage{float}
\usepackage{epstopdf}
\usepackage{pifont}
\usepackage{fixltx2e}
\usepackage{amsmath}
\usepackage{multirow}
\usepackage{url}
\usepackage{verbatim}
\usepackage{caption}
\usepackage{adjustbox}
\usepackage{booktabs}
\usepackage{threeparttable}
\usepackage{makecell}
\usepackage{tabularx}
\usepackage{arydshln}
\usepackage[normalem]{ulem}

\usepackage{algorithm}
\usepackage{ifthen}

\usepackage{algorithmic}

\usepackage{array}
\usepackage{textcomp}
\usepackage{stfloats}
\usepackage{xcolor}
\usepackage{siunitx}
\pdfpxdimen=\dimexpr 1 in/72\relax  %

\newcommand{\eg}{\emph{e}.\emph{g}.}

\newcommand{\cmark}{\ding{51}}%
\newcommand{\xmark}{\ding{55}}%

\newcommand{\jhang}[1]{#1}
\newcommand{\joey}[1]{#1}
\newcommand{\nicholas}[1]{#1}
\newcommand{\rebuttal}[1]{#1}

\newcommand{\final}[1]{#1}

\def\MYTITLE{Toward Deep Representation Learning for Event-Enhanced Visual Autonomous Perception: \\the eAP Dataset}
\def\MYTITLEPDF{Toward Deep Representation Learning for Event-enhanced Visual Autonomous Perception: the eAP Dataset}

\definecolor{eccvblue}{rgb}{0.12,0.49,0.85}

\makeatletter
\let\NAT@parse\undefined
\makeatother
\usepackage[pagebackref=false,breaklinks=true,colorlinks,citecolor=eccvblue,bookmarks=true,bookmarksnumbered=true]{hyperref}
\hypersetup{
  pdftitle={\MYTITLEPDF},
  pdfsubject={Robotics, Computer Vision},
  pdfauthor={Jinghang Li, Shichao Li, Qing Lian, Peiliang Li, Xiaozhi Chen, Yi Zhou},
  pdfkeywords={Deep Learning in Robotics and Automation; Computer Vision for Transportation; Sensor Fusion; Event-based Vision}
}

\usepackage{cleveref}

\title{\MYTITLE}
\author{Jinghang Li$^{\ast}$, Shichao Li$^{\ast}$, Qing Lian, Peiliang Li, Xiaozhi Chen, Yi Zhou$^{\dagger}$%
\thanks{Jinghang Li and Yi Zhou are with the Neuromorphic Automation and Intelligence Lab (NAIL) at School of  AI \& Robotics, Hunan University, Changsha 410082, China. E-mail: {\small \{jhanglee, eeyzhou\}@hnu.edu.cn.}}
\thanks{Shichao Li is with ByteDance, Shenzhen 518063, China. E-mail: shichao.li@bytedance.com}
\thanks{Qing Lian, Peiliang Li, and Xiaozhi Chen are with Zhuoyu Technology, Shenzhen 518055, China. Email: {\small lianqing1997@gmail.com; peiliang.li@zyt.com; xiaozhi.chen@zyt.com}.}
\thanks{$\ast$ denotes equal contribution. Corresponding author ($\dagger$): Yi Zhou.}
}

\definecolor{light-gray}{gray}{0.5}

\begin{document}

\setcounter{figure}{-2} %
\makeatletter

\g@addto@macro\@maketitle{
\vspace{4ex}
  \centering
  \includegraphics[width=0.95\textwidth]{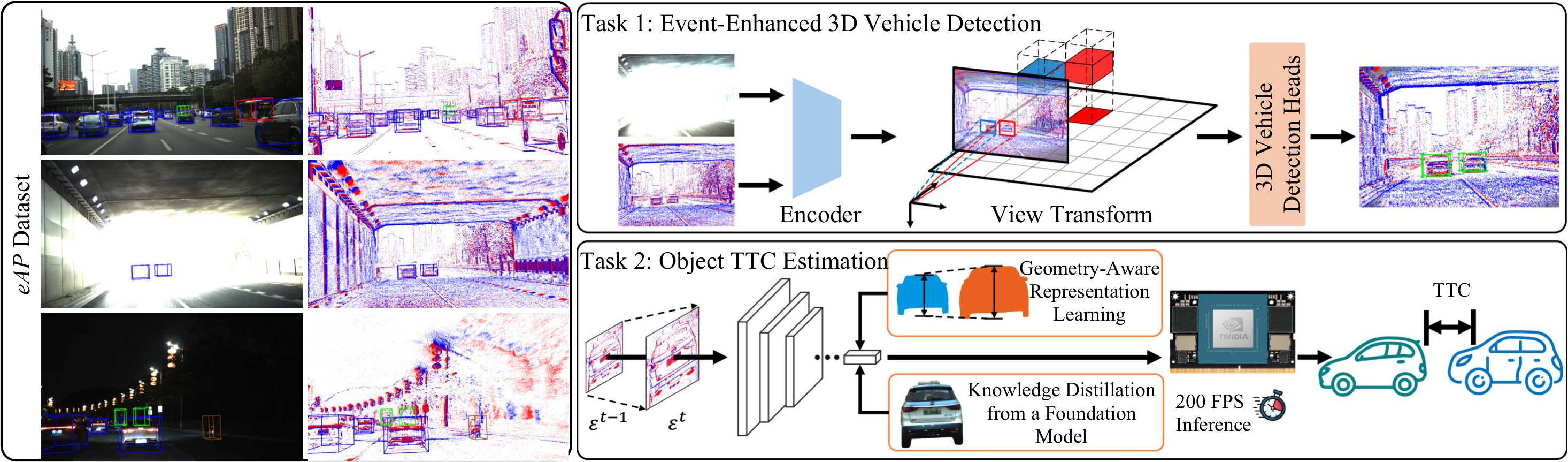}
   \captionof{figure}{
   \rebuttal{Overview of the proposed \emph{eAP} dataset and its use in learning deep representations for two event-enhanced perception tasks.
   (a) Left: Examples from the \emph{eAP} dataset, a large-scale multi-modal dataset featuring event cameras for visual autonomous perception in driving scenarios. 
   (b) Top-right: Framework for event-enhanced 3D vehicle detection via deep representation learning. An encoder processes early-fused event-RGB data, followed by a view transform, enhancing robustness in challenging Standard-Dynamic-Range (SDR) camera conditions.
   (c) Bottom-right: Framework for event-enhanced object Time-to-Collision (TTC) estimation. A novel geometry-aware representation learning approach, employing knowledge distillation from a foundation model, yields a deep model that outperforms previous event-based methods and achieves 200 FPS on an NVIDIA Orin NX.}}
\label{fig:teaser}

\vspace{-2ex}
}
\makeatother
\maketitle

\begin{abstract}
Recent visual autonomous perception systems achieve remarkable performances with deep representation learning. However, they fail in scenarios with challenging illumination. While event cameras can mitigate this problem, there is a lack of a large-scale dataset to develop event-enhanced deep visual perception models in autonomous driving scenes. To address the gap, we present the \emph{eAP} (\textbf{e}vent-enhanced \textbf{A}utonomous \textbf{P}erception) dataset, the largest dataset with event cameras for autonomous perception. We demonstrate how \emph{eAP} can facilitate the study of different autonomous perception tasks, including 3D vehicle detection and object time-to-contact (TTC) estimation, through deep representation learning. Based on \emph{eAP}, we demonstrate the first successful use of events to improve a popular 3D vehicle detection network in challenging illumination scenarios. \emph{eAP} also enables a devoted study of the representation learning problem of object TTC estimation. We show how a geometry-aware representation learning framework leads to the best event-based object TTC estimation network that operates at 200~FPS. The dataset, code, and pre-trained models will be made publicly available for future research. 
\end{abstract}

\section*{Multimedia Material}
\noindent Video: {\small \url{https://youtu.be/6nuFrPViD3U}}\\
Project Page: {\small\url{https://nail-hnu.github.io/eAP_dataset/}}

\section{Introduction}
\label{sec:intro}

\IEEEPARstart{D}{etection} and motion estimation of vehicles in 3D space has indispensable significance for autonomous driving systems to navigate in complex environments. Numerous studies address this problem using LiDAR sensors~\cite{wu2020deep}, yet they incur high costs and suffer from the sparsity of the point cloud. Visual 3D object perception~\cite{ma20233d} using RGB cameras, on the other hand, can fully unleash the potential of low-cost autonomous systems.

\joey{
The performance of frame-based Standard-Dynamic-Range (SDR) cameras, a cornerstone of conventional visual perception systems, is fundamentally limited in challenging conditions. Their reliance on a global exposure time causes motion blur and over/under-exposure in scenes with fast-moving objects or extreme lighting. 
Neuromorphic event cameras overcome these limitations by asynchronously measuring per-pixel brightness changes~\cite{gallego2020event,gehrig2024low,chakravarthi2025recent}, providing high temporal resolution and High-Dynamic-Range (HDR). 
This work explores how these distinctive properties can be leveraged to enhance the robustness and safety of autonomous perception in diverse outdoor driving scenarios.}

Specifically, we study two important sub-tasks for event-enhanced autonomous perception: 3D vehicle detection and object time-to-contact (TTC) estimation. The former task aims to discover and locate vehicles in 3D space, and the latter performs object motion estimation to prevent collisions. This selection of tasks has three reasons:
\begin{itemize}
    \item \emph{Importance of tasks:} These two tasks span from object detection~\cite{zou2023object} to motion estimation~\cite{barfoot2024state}, which are both fundamental problems in robotic perception.
    \item \emph{Scope of representations:} Recent autonomous perception approaches are dominated by two types of representations: the bird's eye view (BEV) representation~\cite{ma2024vision} and the classical front-view (FV) representation~\cite{zhao2019object,luo2021multiple,minaee2021image}. The former builds dense voxels in the 3D space for high accuracy, while the latter utilizes image-level 2D representations for fast inference. Our choice of problems covers these two mainstream approaches.
    \item \emph{Value of applications:} Studying these two tasks enables an autonomous agent to identify vehicles and mitigate potential collisions. FV TTC estimation systems are widely adopted in modern \emph{Automated Emergency Braking} (AEB) systems, which include the \emph{Forward Collision Warning} (FCW) functionality~\cite{fildes2015effectiveness,cicchino2017effectiveness} for fast response to potential collisions in life-critical situations.   
\end{itemize}
Unfortunately, we encounter a data dilemma when pursuing performance beyond the state of the art.
\vspace{1ex}
\begin{table*}[t!]
\begin{threeparttable} 
\centering
\setlength{\tabcolsep}{3pt} %
\resizebox{0.98\linewidth}{!}{%
\normalfont
\begin{tabular}
{lcccccccccc}
\toprule
 \makecell*[l]{Datasets}
&\makecell*[c]{Year}
&\makecell*[c]{Duration\\ (in hours)} %
&\makecell*[c]{\#3D Bbox}
&\makecell*[c]{TTC\\ Anno}
&\makecell*[c]{\#Class}
&\makecell*[c]{\#Resolution\\ (RGB)}
&\makecell*[c]{\#Resolution\\ (Event)}
&\makecell*[c]{Aim}
&\makecell*[c]{Data\\ source}
&\makecell*[c]{Scenarios} \\ 
\midrule %

    MVSEC~\cite{MVSEC} &
    2018 &
    0.5 &
    - &
    - &
    - &
    $752\times480$ &
    $346\times260$ &
    Eval & Collected & Urban \\
    
    DDD20~\cite{DDD2000} & 
    2020 & 
    51 & 
    - & 
    - & 
    - & 
    $346\times260$ & 
    $346\times260$ & 
    Train\&Eval & Collected &  Expressways, Highways, Urban \\
    
    DSEC-Det~\cite{gehrig2024low} & 
    2021 & 
    1 & 
    - & 
    - & 
    6 & 
    $1440\times1080$ & 
    $640\times480$ & 
    Train\&Eval & 
    Collected & 
    Urban, Suburban, Rural areas \\

    \rebuttal{VECtor}~\cite{gao2022vector} & 
    \rebuttal{2022} & 
    \rebuttal{0.36} & 
    \rebuttal{-}& 
    \rebuttal{-}&  
    \rebuttal{-}&  
    \rebuttal{$1224\times1024$} & 
    \rebuttal{$640\times480$} & 
    \rebuttal{Eval} & 
    \rebuttal{Collected} &
    \rebuttal{Indoor, School} \\
    
    M3ED~\cite{M3ED} & 
    2023 & 
    \rebuttal{3.42 (car 2)\tnote{1}} & 
    - & 
    - & 
    11 & 
    $1280\times800$ & 
    $1280\times720$ & 
    Train\&Eval & 
    Collected & 
    Urban, Forest \\ 
    
    \rebuttal{DSEC-3DOD}~\cite{cho2025ev}\rebuttal{\tnote{2}} & 
    \rebuttal{2025} & 
    \rebuttal{0.15} & 
    \rebuttal{\checkmark}& 
    \rebuttal{-}&  
    \rebuttal{2}&  
    \rebuttal{$1440\times1080$} & 
    \rebuttal{$640\times480$} & 
    \rebuttal{Eval} & 
    \rebuttal{Collected} &
    \rebuttal{Urban, Suburban, Highways} \\
    
    \midrule
    
    Crash to Not Crash~\cite{kim2019crash} & 
    2019 & 
    - & 
    - & 
    - & 
    1 & 
    $710\times400$ & 
    - & 
    Eval & 
    Internet & 
    - \\
    
    DADA-2000~\cite{DADA2000} & 
    2019 & 
    - & 
    - & 
    - & 
    7 & 
    $1584\times660$ & 
    - & 
    Train\&Eval & 
    Internet & 
    - \\
    
    TSTTC~\cite{TSTTC} & 
    2023 & 
    27.8 & 
    - & 
    \checkmark & 
    1 & 
    $1024\times576$ & 
    - & 
    Train\&Eval & 
    Collected & 
    Urban, Highway \\
    
    EvTTC~\cite{sun2024evttc} & 
    2024 & 
    0.2 & 
    - & 
    \checkmark & 
    2 &
    $1920\times1200$ & 
    $1280\times720$ & 
    Eval & 
    Collected & 
    Test track, Clear Road \\

    \rebuttal{FCWD}~\cite{li2024strttc} & 
    \rebuttal{2024} & 
    \rebuttal{0.05} & 
    \rebuttal{-} & 
    \rebuttal{\checkmark} & 
    \rebuttal{1} & 
    \rebuttal{$1920\times1200$} & 
    \rebuttal{$1280\times720$} & 
    \rebuttal{Eval} & 
    \rebuttal{Collected} & 
    \rebuttal{Clear Road}
     \\
    
    \rebuttal{EV-TTC}~\cite{bisulco2025ev}\rebuttal{\tnote{3}} & 
     \rebuttal{2025} & 
    \rebuttal{3.42 (car 2)\tnote{1}} & 
    \rebuttal{-} & 
    \rebuttal{\checkmark} & 
    \rebuttal{11} & 
    \rebuttal{$1280\times800$} & 
    \rebuttal{$1280\times720$} & 
    \rebuttal{Train\&Eval} & 
    \rebuttal{Collected} & 
    \rebuttal{Urban, Forest}
     \\
    \midrule
    
    \emph{eAP} (Ours) & 
    2025 & 
    4.8 & 
    532K+ & 
    \checkmark &
    \rebuttal{8} & 
    $1920\times1200$ & 
    $1280\times720$ & 
    Train\&Eval & 
    Collected & 
    Highways, Urban, Low-light \\ 
    \bottomrule
    \end{tabular}
}
\begin{tablenotes}
    \item[1] \rebuttal{The M3ED dataset~\cite{M3ED} includes 1.9 hours of driving data (approximated to 2 hours) out of a 3.42-hour total.}
    \item[2] \rebuttal{The DSEC-3DOD~\cite{cho2025ev} is built upon the DSEC~\cite{gehrig2024low}, providing novel 3D bounding box annotations for a portion of its ``zurich city'' sequence.}
    \item[3] \rebuttal{The EV-TTC dataset~\cite{bisulco2025ev} is built upon the M3ED dataset~\cite{M3ED}, which is the first event dataset offering dense TTC ground truth.}
\end{tablenotes}

\caption{A comprehensive overview of recent collision prediction and TTC estimation datasets. 
\emph{eAP} features a large dataset scale, diverse scenes, high resolution, \final{a narrow-baseline setup for cross-modal tasks}, along with object-level 3D bounding box and TTC annotations.
}
\label{tab:related_datasets}
\end{threeparttable} 
\end{table*}

\vspace{1ex}

The potential of event cameras for visual 3D vehicle detection remains largely unexplored.
Recent approaches adopt an end-to-end
BEV paradigm and have achieved remarkable performance~\cite{ma2024vision}. 
However, they require a large amount of training data to generalize in complex environments. 
To facilitate studying the problem of event-enhanced visual 3D vehicle detection, a large-scale dataset with event cameras and 3D box labels is highly desirable. 
Unfortunately, to our best knowledge, no such dataset exists. 

A similar dilemma exists for the event-enhanced visual TTC estimation problem. TTC is defined as the time remaining until the object's depth reaches zero, assuming the current relative motion between the camera and the object remains unchanged~\cite{horn2007time}. Previous event-based TTC estimation methods employ hand-crafted features to establish data association across events.
These methods~\cite{gallego2018unifying,dinaux2021faith,dupeyroux2021neuromorphic,shiba2023fast,nunes2023time,li2024strttc} typically formulate the TTC estimation problem as a geometric-model-fitting task.
However, these methods often employ overly simplified parametric models and are evaluated in environments with limited variation in object types, lighting conditions, and TTC ground truths. 
As a result, they may not adequately capture the complexity and diversity of real-world autonomous driving scenarios. More importantly, these model-fitting-based approaches cannot learn data-driven representations, which prevents them from leveraging the growing volume of event data to improve performance and generalization capabilities.
\jhang{Large-scale event-camera datasets for autonomous perception that offer sufficient scene diversity and object-level TTC labels are currently scarce},
hindering the development of deep learning-based approaches that typically require extensive data to achieve robust performance.

\joey{To address this dataset shortage and advance learning-based, event-enhanced 3D object perception, we introduce the \emph{eAP} (\textbf{e}vent-enhanced \textbf{A}utonomous \textbf{P}erception) dataset.}
The \emph{eAP} dataset encompasses a diverse range of scenes, objects, and corresponding 3D bounding box and TTC ground truths, enabling a comprehensive study of deep representation learning towards enhancing visual autonomous perception with events.

\joey{Leveraging the \emph{eAP} dataset, we present the first in-depth study of event-enhanced 3D vehicle detection and object TTC estimation via the framework of deep representation learning~\cite{bengio2013representation}.}
Specifically, we propose the first deep-learning-based 3D vehicle detection approach 
that fuses RGB and event (RGB-E) data.
We also propose \emph{Garl-TTC} (\textbf{G}eometry-\textbf{A}ware \textbf{R}epresentation \textbf{L}earning for object TTC estimation), a novel framework for event-enhanced visual object TTC estimation. 
\emph{Garl-TTC} considers the geometry of TTC, which leads to a model design that uses a visible object height representation for TTC estimation. 
To further enhance representation learning, we guide the model to learn foreground-aware representations by distilling knowledge from a foundation model. 
Our framework, as shown in Fig.~\ref{fig:teaser}, ensures accurate and reliable 3D vehicle detection and TTC estimation with a single forward pass given RGB-E input.

In summary, the key contributions of this paper are as follows:
\begin{itemize}
    \item We introduce a new curated dataset \emph{eAP}, collected using multi-modal sensors, that captures a wide range of driving scenes, including varying lighting conditions, object types, and motion dynamics, specifically tailored for advancing
    research in event-enhanced visual autonomous perception.
    \item We present the first study of the event-enhanced visual 3D vehicle detection problem and demonstrate how event cameras can improve the state-of-the-art visual 3D vehicle models in challenging illumination scenarios.
    \item We present the first dedicated framework (\emph{Garl-TTC}) that addresses object-level TTC estimation through geometry-aware deep representation learning with event-based cameras. Extensive evaluation and benchmarking demonstrate that our method achieves state-of-the-art accuracy and exhibits significantly improved generalization capability over existing methods.
\end{itemize}

\joey{
The remainder of this paper is structured as follows. 
We review related work in Sec.~\ref{sec:related} and introduce the \emph{eAP} dataset in Sec.~\ref{sec:dataset}. 
After formulating the core perception problems in Sec.~\ref{sec:formulation}, we present our novel event-enhanced 3D vehicle detection model in Sec.~\ref{sec:method_3dod}. 
Sec.~\ref{sec:method_ttc} then details our framework for object TTC estimation, including a baseline and our proposed geometry-aware representation. 
Finally, Sec.~\ref{sec:experiment} provides extensive experiments, ablation studies, and on-board deployment results.}

\section{Related Work}
\label{sec:related}

We provide a comprehensive overview of prior research, including relevant event-camera datasets, visual 3D vehicle detection, and visual object TTC estimation. 

\noindent\textbf{Datasets for event-enhanced perception.}
Over the past decade, numerous datasets have been developed for perception tasks such as TTC estimation and collision recognition in autonomous driving, as summarized in Table~\ref{tab:related_datasets}. 
Datasets such as DADA-2000~\cite{DADA2000} and CrashDataset~\cite{kim2019crash} compile videos of emergency braking and collision scenarios sourced from the Internet. 
However, they lack continuous TTC labels, limiting their utility for precise TTC estimation. 
TSTTC~\cite{TSTTC} addresses this limitation by providing 2D bounding boxes and TTC ground-truth values, but it does not include event sensor data.
Existing event camera datasets, such as MVSEC~\cite{MVSEC}, are constrained by relatively low spatial resolution (346×260 pixels). 
\jhang{While recent datasets like DSEC~\cite{DSEC} improve sensor resolution, they lack ground-truth object TTC annotations and face challenges in accurately labeling distant objects due to the limited resolution of their 16-beam LiDARs.}
\jhang{The M3ED~\cite{M3ED} dataset provides dense point clouds 
using an OS1-64 LiDAR in various scenes. However, it is not 
specialized for autonomous driving applications, as only part 
of the data was collected in a moving vehicle. 
The EV-TTC~\cite{bisulco2025ev} dataset later leveraged M3ED to generate dense (i.e., per-pixel) TTC annotations. 
\final{While dense TTC provides valuable per-pixel motion information, object-level TTC
offers a complementary paradigm for object-centric autonomous perception.}
Existing datasets like EvTTC~\cite{sun2024evttc} offer object-level TTC labels, but their scope is limited to 32 specific EuroNCAP~\cite{EuroNCAP2024} test scenarios. 
These scenarios primarily consist of car-following situations on straight roads and lack the scene diversity required for training and evaluating robust, real-world systems.}

In contrast, we propose \emph{eAP}, the first large-scale event camera dataset \nicholas{with rich object-level labels dedicated to autonomous perception.}

\noindent\rebuttal{\textbf{Event-enhanced 3D vehicle detection.} 
Numerous previous studies tackle the visual 3D vehicle detection problem~\cite{mao20233d}. Early studies follow a 2D object detection paradigm~\cite{zou2023object} and perform anchor-based~\cite{brazil2019m3d,ding2020learning} or anchor-free~\cite{liu2020smoke} classification and regression on FV images. Recent studies adopt a BEV paradigm~\cite{ma2024vision} and learn deep representations on a BEV feature map for object detection, achieving state-of-the-art performance. However, they share the drawbacks of frame-based perception with SDR cameras and fail in situations with extreme illumination. It is tempting to incorporate the advantages of event cameras for multi-modal perception, yet it is intimidating to collect and label a massive amount of data to train a data-hungry BEV object detector. Our study takes the first step to address the data bottleneck and presents the first event-enhanced visual 3D vehicle detection model. A concurrent study, Ev-3DOD~\cite{cho2025ev}, also utilizes events for 3D object detection, yet focuses on utilizing LiDAR sensors while our study employs cost-effective cameras.}

\noindent\textbf{Frame-based TTC estimation}.
Existing research on TTC estimation from frames can be broadly categorized into optimization-based and learning-based approaches.  
Optimization-based approaches~\cite{souhila2007optical, stabinger2016monocular, poiesi2016detection} rely on parametric models that describe the relationship between TTC and visual cues such as \nicholas{optical flow~\cite{nelson1989obstacle, yang2020upgrading}}, normal flow~\cite{meyer1992estimation}, and scale variation of features~\cite{dagan2004forward, negre2008real}. 
These methods formulate TTC inference as an optimization problem, which requires careful initialization, hyper-parameter tuning for data pre-processing, and iterative updates. 
Moreover, these methods are often limited in their ability to discover generalizable representations as the volume of collected data grows.  
Learning-based approaches~\cite{manglik2019forecasting, TSTTC, badki2021binary}, on the other hand, directly estimate TTC using a sequence of frames as input. 
For instance, BinaryTTC~\cite{badki2021binary} formulates the TTC estimation problem as a pixel-level classification task, where different TTC thresholds are used to classify pixels. Despite the significant progress made by frame-based methods, they are inherently limited by the high latency of frame-based cameras. \rebuttal{In addition, frame-based methods that use SDR cameras often perform poorly in scenes with challenging illumination.} In contrast, event-based cameras offer distinct advantages in such scenarios.
\vspace{1ex}
\begin{figure}[t!] 
    \centering
    \includegraphics[width=0.47\textwidth]{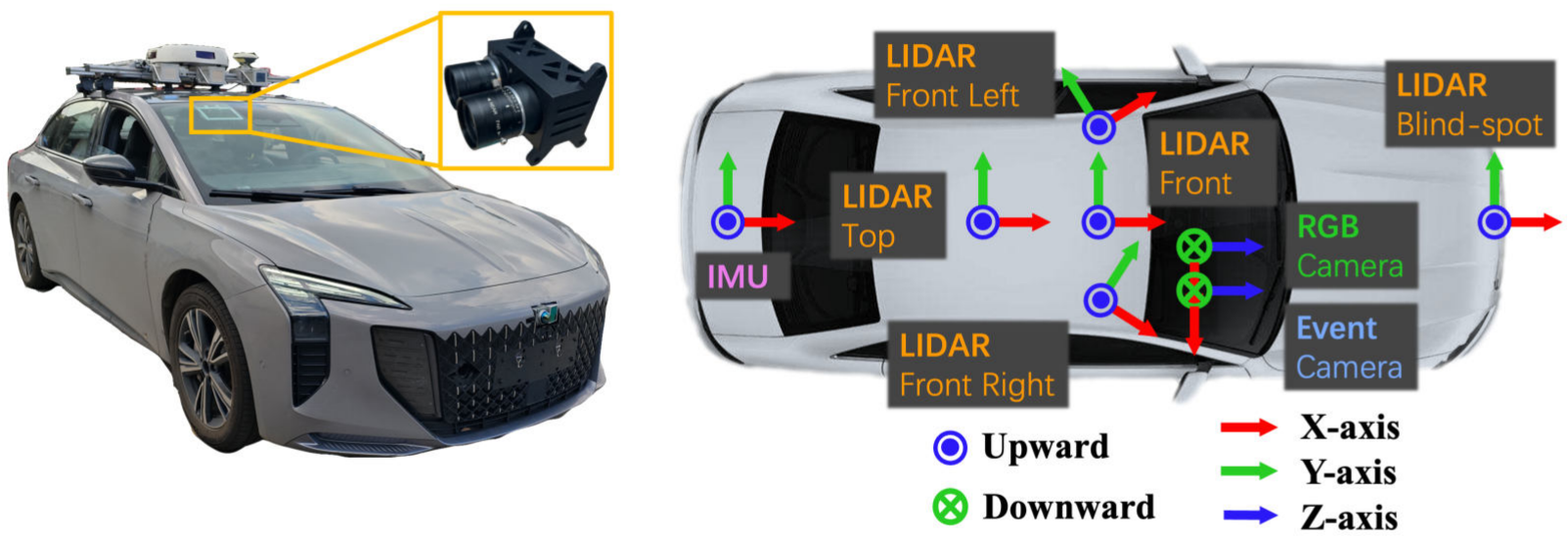}
    \caption{
    \rebuttal{Sensor configuration for the \emph{eAP} dataset. 
    The event camera and RGB camera are rigidly mounted with a narrow baseline of 3 cm, as highlighted in the zoomed-in image. 
    The positions and coordinate systems of the sensors are depicted in the bird's eye view.}}
    
    \label{fig:sensor_setup}
    
    \vspace{2ex}
    
    \setlength{\tabcolsep}{3pt}
    \resizebox{0.49\textwidth}{!}{
    \begin{tabular}{cll}
        \toprule
         Sensor Type & Description & Acquisition Settings \\
        \midrule
        \makecell*[c]{FLIR Blackfly S} & 
        \makecell[l]{Resolution: 1920×1200 \\
                    Optical format: 1/2.3'' \\
                    Pixel size: $3.45\,\mu\text{m} \times 3.45\,\mu\text{m}$ \\ 
                    FoV: $45.4^{\circ} \times 29.3^{\circ}$} & 
        \makecell[l]{
            Capture Frequency: 10Hz \\
            Target Grey: \\
            \quad mode: fixed \\
            \quad value: 45\% \\
            Auto Exposure: \\
            \quad mode: continuous \\
            \quad metering mode: partial \\
            \quad exposure time limit: 1-20ms \\
            \quad gain limit: 0-30db \\
            \quad damping: 0.1 \\
            \quad control priority: exposure time \\
            White Balance: \\
            \quad mode: auto \\
            \quad profile: outdoor \\
            \quad limit: 0.5 - 4 \\
            \quad damping: 0.1
            } 
            \\
        \midrule 
        \makecell*[c]{Prophesee EVK4} & 
        \makecell[l]{Resolution: 1280×720 \\
                    Optical format: 1/2.5'' \\
                    Pixel size: $4.86\,\mu\text{m} \times 4.86\,\mu\text{m}$\\
                    FoV: $42.5^{\circ} \times 24.7^{\circ}$} &
        \makecell[l]{
        bias diff on: 20\\
        bias diff off: 20\\
        bias fo: 0\\
        bias hpf: 0\\
        bias refr: 0}
        \\
        \midrule 
        \makecell*[c]{3$\times$Livox TELE-15} & 
        \makecell[l]{Range: 320m@10\% NIST \\ 
                    Accuracy: $(1\sigma@20\text{m}) \pm 2\text{cm}$ \\
                    FoV: $14.5^{\circ} \times 16.2^{\circ}$ \\
                    Point Rate: 240,000 points/s} &
        \makecell[l]{Capture Frequency: 10Hz}
                    \\
        \midrule 
        \makecell*[c]{1$\times$Livox Mid360} & 
        \makecell[l]{Range: 40m@10\% NIST \\ 
                    Accuracy: $(1\sigma@10\text{m}) \pm 2\text{cm}$ \\ 
                    FoV: Horizontal: $360^{\circ}$ \\
                    ~~~~~~~Vertical: $-7^{\circ} \sim 52^{\circ}$ \\
                    Point Rate: 200,000 points/s} &
        \makecell[l]{Capture Frequency: 10Hz}
                    \\
        \midrule 
        \makecell*[c]{u-blox ZED-F9K} & 
        \makecell[l]{GNSS: GPS+GLO+GAL+BDS \\
                    Velocity accuracy: 0.05 m/s \\
                    Pos Accuracy: RTK 0.2 m} &
        \makecell[l]{Capture frequency: 10Hz}
                    \\
        \bottomrule 
    \end{tabular}
    }
    
    \captionof{table}{\rebuttal{Details of the sensors used in the data collection vehicle.}}
    \label{tab:sensor_spec}
\end{figure}

\noindent\textbf{Event-enhanced TTC estimation}. 
Event-based TTC estimation has attracted significant attention in both robotics~\cite{Sanket20icra,falanga2020dynamic,walters2021evreflex,rodriguez2022free} and neuromorphic vision~\cite{clady2014asynchronous}. 
Recent advancements in event-based solutions have shifted toward flow-free methodologies, which bypass explicit optical flow computation by leveraging geometric model fitting. 
For instance, Contrast Maximization (CMax)~\cite{gallego2018unifying} directly optimizes parametric motion models from event data. 
Subsequent studies~\cite{shiba2023fast,mcleod2023globally} enhance CMax, improving its efficiency and stability. 
Building on this model-fitting paradigm, ETTCM~\cite{nunes2023time} integrates motion and depth estimation to generate time-to-collision maps for navigation tasks. 
Further progress is achieved by Li et al.~\cite{li2024strttc}, who introduce a two-stage approach combining a time-variant affine solver with spatiotemporal refinement.
Despite these advancements, existing methods are constrained by their reliance on oversimplified geometric assumptions (\eg, predefined motion patterns) and iterative optimization processes. 
While effective for simple motion scenarios, they often struggle in complex real-world environments characterized by diverse object motion patterns and event noise induced by challenging illumination conditions. 
\nicholas{To overcome these limitations, deep learning-based approaches are promising due to their ability to learn robust deep representations directly from event data. This ability overcomes the fragile, hand-crafted models of classical approaches and offers superior performance in complex scenarios.}
\jhang{A concurrent study, EV-TTC~\cite{bisulco2025ev}, exemplifies this learning-based trend. It introduces a system for dense TTC estimation, emphasizing a hardware-aware, multi-temporal scale representation designed for real-time performance on resource-constrained mobile robots.}
In contrast to dense TTC estimation, our work proposes the first geometry-aware deep learning framework for direct object-level TTC estimation, which is \final{highly relevant for many object-centric autonomous perception applications.} 
Our method builds upon the geometry of TTC estimation as in classical studies~\cite{nelson1989obstacle}, yet achieves end-to-end estimation without explicitly computing optical flow. This approach eliminates the need for optical flow ground truth or scene structure priors, enabling robust performance in dynamic environments.

\section{The \emph{eAP} Dataset}
\label{sec:dataset}

This section describes the \emph{eAP} dataset and compares it with previous event-based datasets. We detail the data collection and annotation process and show the dataset statistics. 

\subsection{Sensor setup}
Our data collection vehicle, as illustrated in~Fig.~\ref{fig:sensor_setup}, is equipped with a sensor suite consisting of LiDAR sensors, an RGB camera, and an event camera. Compared to the DSEC dataset~\cite{DSEC}, which utilizes a 640×480 event camera and a 16-beam LiDAR, our system features two significant hardware upgrades: (1) a 1280×720 event camera capable of capturing finer visual details and (2) five forward-facing Livox LiDAR sensors that can capture dense point cloud for accurate labeling of vehicles. \rebuttal{Detailed parameters of the sensors and the settings during data collection are summarized in Tab.~\ref{tab:sensor_spec}.}

\nicholas{Before data collection, we meticulously calibrate the RGB camera's Image Signal Processor (ISP) parameters\footnote{\url{https://github.com/ros-drivers/flir_camera_driver}} in a standard lightbox environment to ensure appropriate imaging quality. We first calibrate the White Balance parameter and then set the Target Grey parameter to 45\%. Both Auto Exposure (AE) and Auto White Balance (AWB) are enabled to adapt to real-world illumination. Despite these settings, an SDR camera is limited to a single global exposure time and global gain applied across all pixels, which forces an inevitable tradeoff when capturing HDR scenes. As illustrated in Fig.~\ref{fig:teaser}, when the AE algorithm uses a longer exposure or higher gain to properly expose dark regions (e.g., tunnels), bright regions (e.g., daylight outside) are inevitably overexposed. Conversely, properly exposing highlights risks underexposing the dark regions. During data collection, we choose a balanced configuration by setting the Control Priority parameter to Exposure Time Priority and the Limits parameter to 1-20ms and 0-30dB. We find that this configuration is appropriate for an autonomous driving dataset. It prioritizes minimal motion blur by optimizing exposure time first, before applying gain to boost brightness, thereby balancing overall scene brightness against motion artifacts and noise.}

\nicholas{Bias is a critical parameter of the event camera that controls the sensor's sensitivity and the event noise floor. We follow the official Prophesee documentation~\footnote{\url{https://docs.prophesee.ai/stable/hw/manuals/biases.html\#bias-tuning}} to tune the bias\_diff\_on and bias\_diff\_off parameters, which manage the sensitivity threshold for positive and negative contrast changes. We perform experimental tuning using Metavision Studio under varied lighting conditions in representative daytime and nighttime scenes. We tune the parameters to strike a balance that suppresses the majority of background noise events without excessively sacrificing sensitivity to real-world dynamics. We keep the same bias parameters during our data collection to achieve consistency across all data sequences.}

\rebuttal{We estimate the vehicle's high-precision ego-pose in the world frame using the tightly-coupled GNSS-Visual-Inertial fusion system (GVINS) presented in~\cite{cao2022gvins}. This system fuses visual data from our onboard cameras with GNSS and IMU measurements from an \mbox{u-blox} ZED-F9K unit (specifications listed in Tab.~\ref{tab:sensor_spec}).}

\subsection{Synchronization and calibration}
\noindent \textbf{Synchronization.} We implement hardware time synchronization across all sensors to ensure sub-microsecond temporal alignment. This is particularly critical for our dataset since our bounding boxes are labeled on captured point clouds, and many driving scenes include high-speed vehicles. A small temporal synchronization error can result in a large re-projection error of a 3D bounding box on the event camera plane. The LiDAR system achieves microsecond-level synchronization with the data acquisition computer via the IEEE Precision Timing Protocol (PTP)~\cite{eidson2002ieee}. 
A 10 Hz synchronization signal from the industrial-grade control unit is simultaneously broadcast to trigger RGB camera exposures and generate \rebuttal{\emph{External Trigger} within the raw event stream\footnote{\url{https://docs.prophesee.ai/stable/hw/manuals/synchronization.html}}. During the offline post-processing stage, we first parse the raw event stream to extract the microsecond-level timestamps of these \emph{External Trigger}. Since the LiDAR data and RGB frames are captured with the same synchronization signal, we can remap LiDAR and RGB frames onto the event camera's high-precision master clock, achieving accurate offline synchronization.} This synchronization protocol, combined with \rebuttal{the} post-processing \rebuttal{stage}, guarantees system-wide temporal coherence at the microsecond level.

\begin{figure}[t!]
    \centering
    \includegraphics[width=0.47\textwidth]{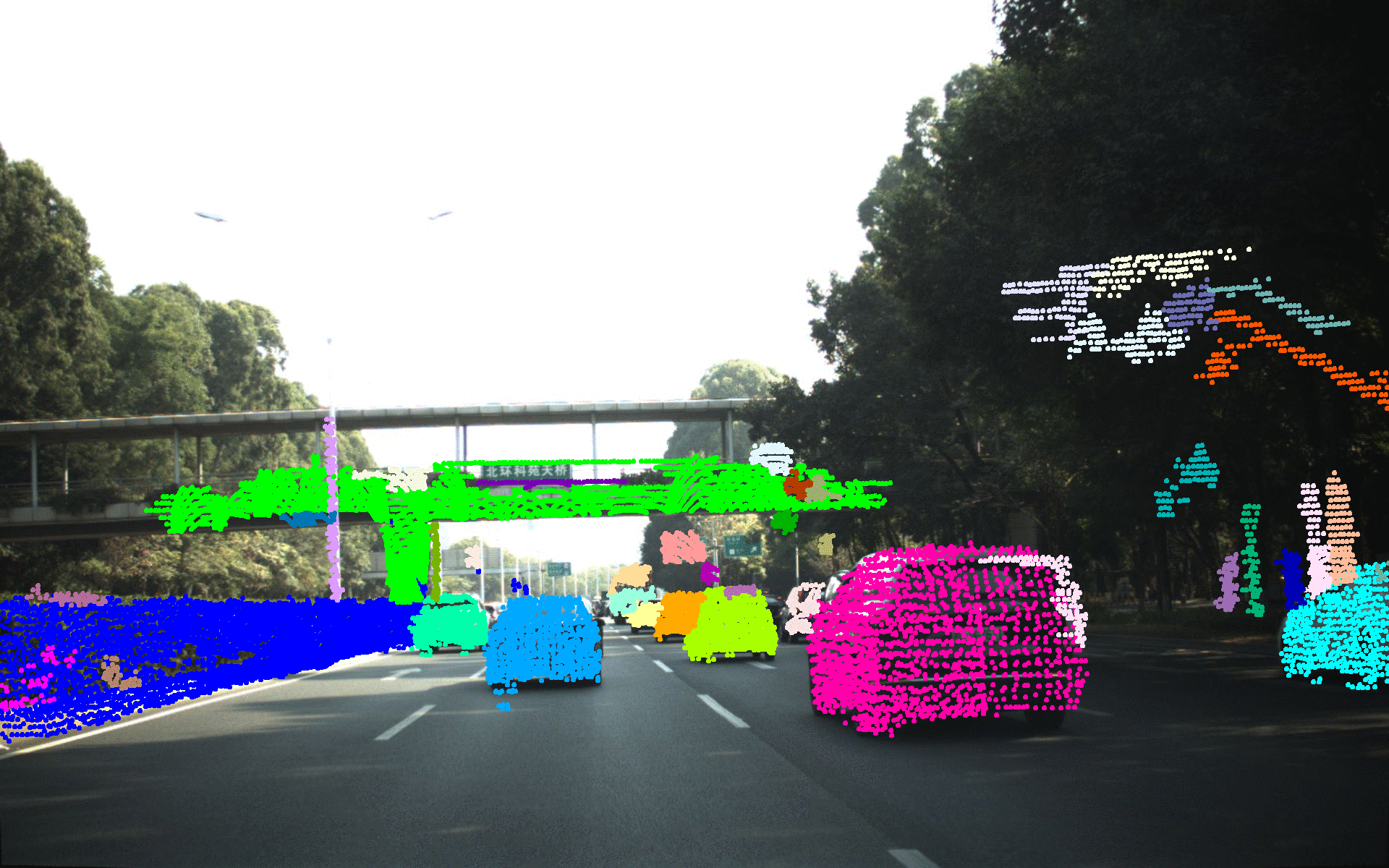}
    \caption{Projected point cloud on the RGB image after using Calib-Anything~\cite{luo2023calib} for calibrating the extrinsic parameters between the LiDAR and the RGB camera. \rebuttal{Different colors represent the distinct clusters identified during the clustering process in~\cite{luo2023calib}. The ground points are omitted for clearer visualization.}}
    \label{fig:calib_anything}
    \vspace{4ex}

    \centering
    \includegraphics[width=0.47\textwidth]{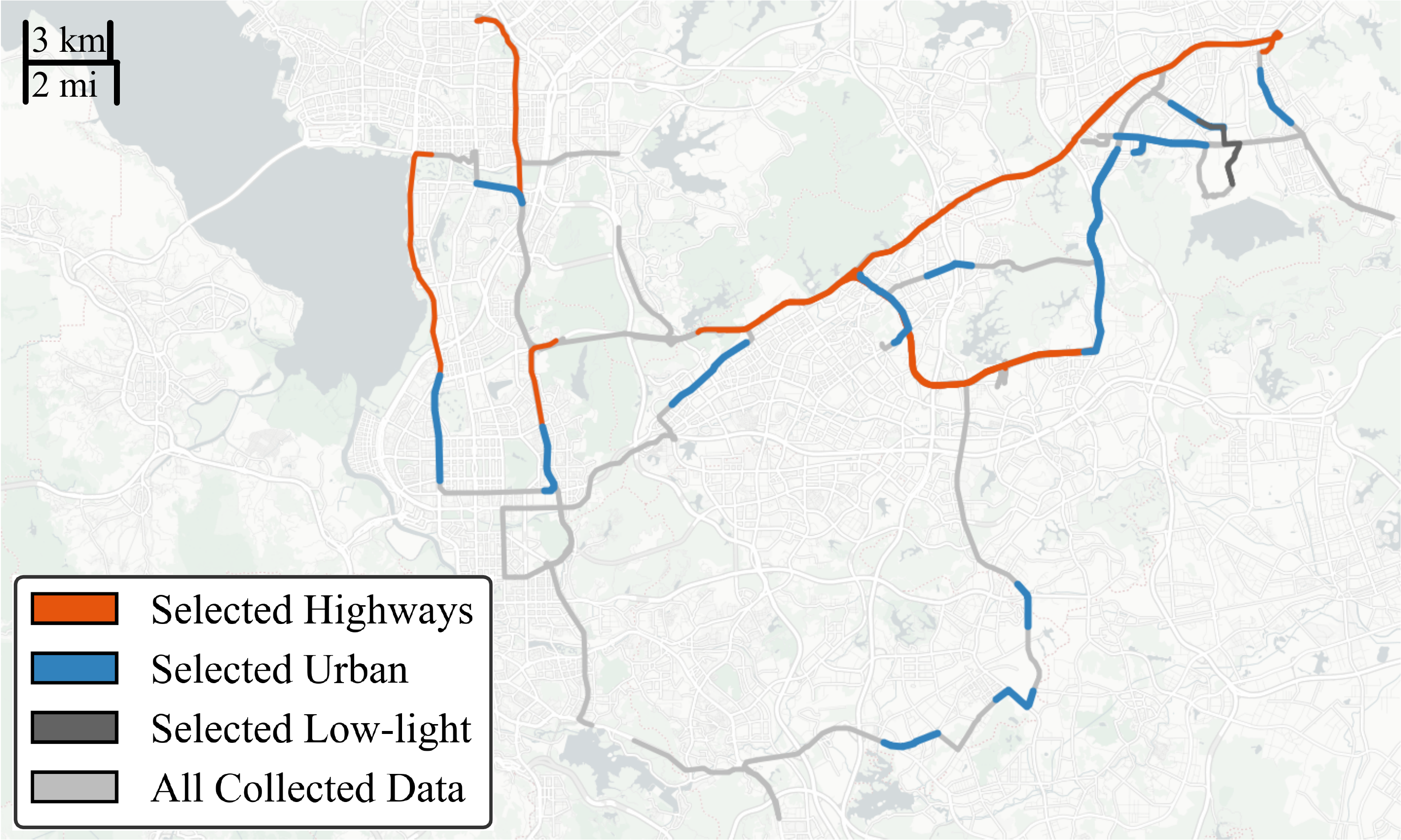}
    \caption{
    \rebuttal{Data collection routes for the \emph{eAP} dataset. 
    The complete raw data (gray) is refined to a 4.8-hour subset (highlighted in color), comprising: a 3-hour highway sequence (178.44 km), a 1h40m urban sequence (52.61 km), and a 10-minute low-light sequence (5.01 km).}}
    \label{fig:data_map}
\end{figure}

\noindent \textbf{Calibration.} The sensor calibration process consists of three components: intrinsic and extrinsic calibration of the RGB-event camera pair, static calibration of the LiDAR units, and cross-modal extrinsic calibration between the LiDAR sensors and the cameras. For the RGB-event camera pair, intrinsic parameters of individual cameras and their inter-sensor extrinsic relationships are determined using the Kalibr toolbox~\cite{furgale2013unified}. Time-synchronized calibration sequences are recorded using a 12$\times$8 checkerboard target. While the RGB camera's intrinsic calibration follows standard procedures, the event camera requires specialized preprocessing: event streams are converted into reconstructed intensity images via the Simple Image Reconstruction library\footnote{\scriptsize \url{https://github.com/berndpfrommer/simple_image_recon}} prior to using the Kalibr toolbox for calibration.
\nicholas{We achieve a high calibration quality of our RGB and event cameras, as validated by the measured low calibration errors. Regarding the intrinsic calibration errors measured by the reprojection error, the RGB camera achieves a standard deviation of $0.303 \text{ px}$ (x-axis) and $0.128 \text{ px}$ (y-axis), while the event camera yields $0.482 \text{ px}$ (x-axis) and $0.221 \text{ px}$ (y-axis). As for the extrinsic calibration error between the sensors, the rotational uncertainty (standard deviation) is less than $0.051^{\circ}$, and the translational uncertainty is confirmed to be sub-millimeter, with standard deviations of $0.16 \text{ mm}$ (x), $0.17 \text{ mm}$ (y), and $0.92 \text{ mm}$ (z).}
The Calib-Anything framework~\cite{luo2023calib} is employed for extrinsic calibration between the LiDAR sensors and the RGB camera. As shown in Fig.~\ref{fig:calib_anything}, the projected LiDAR point cloud (color) aligns well with the image segmentation masks across objects in complex urban driving scenarios. 
\final{Furthermore, to correct motion-induced skewing from the LiDAR sensors' non-repetitive scanning, we perform point-wise deskewing via trajectory interpolation using high-frequency INS and wheel odometry~\cite{xu2022fast}.}

\noindent \final{\textbf{Event-to-RGB mapping.}} 
\final{Given the narrow (3~cm) extrinsic baseline, we map the RGB images to the event camera's view using an ``infinite depth mapping'' as adopted in datasets such as DSEC-Det~\cite{gehrig2024low} and VECtor~\cite{gao2022vector}.} 
This process remaps the RGB image directly onto the event camera's distorted image plane. We first undistort the RGB image, rotate it to match the event camera's orientation, and then re-project it using the event camera's specific distortion model. 
\final{As noted in previous studies, this infinite depth mapping does not guarantee exact pixel-to-pixel correspondence because the transformation is depth-dependent when a baseline exists. However, our narrow baseline limits the maximal disparity to less than 5 pixels across our dataset. Such a minor misalignment is generally considered acceptable for deep learning-based sensor fusion in autonomous perception tasks like object detection.}

\subsection{Data collection}

\begin{figure*}[t]
    \centering
    \includegraphics[width=0.95\textwidth]{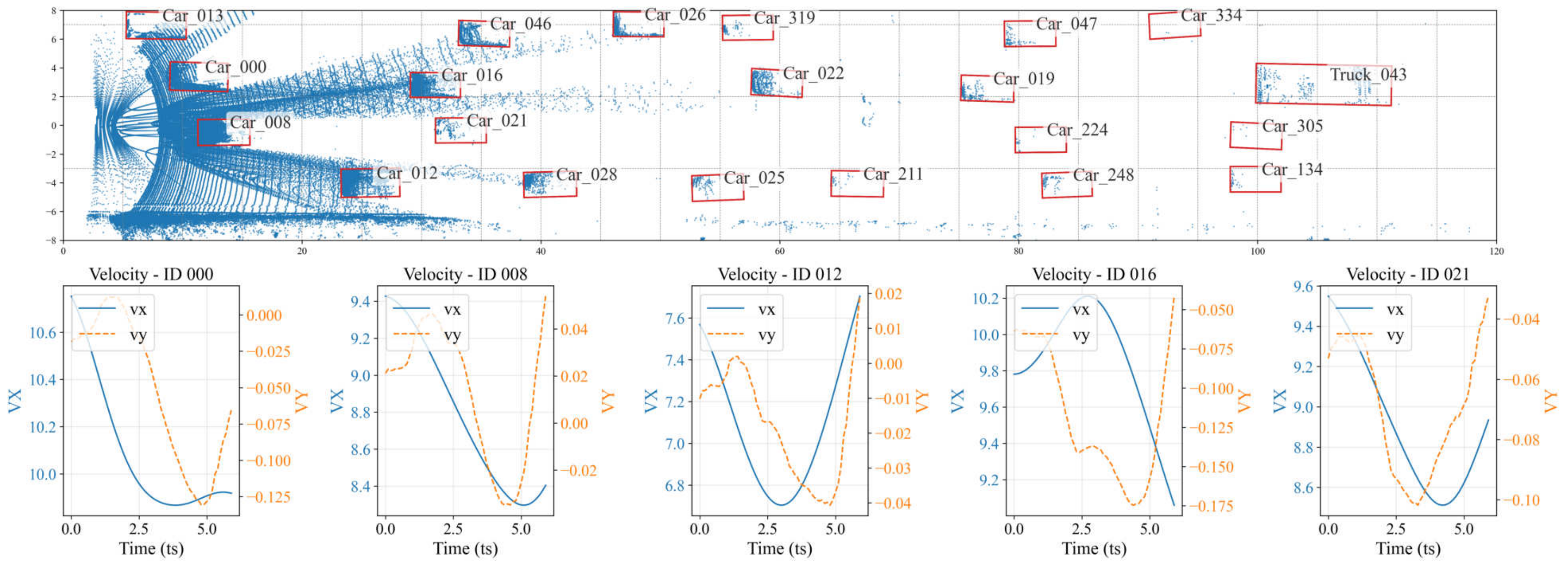}
    \caption{(Top) Bird's eye view (BEV) visualization of the LiDAR point cloud with annotated 3D bounding boxes demonstrating the 3D annotation precision; (Bottom) Plot of velocity curves of exemplar object trajectories.}
    \label{fig:lidar_bev}
    \vspace{-2ex}
\end{figure*}

We design a comprehensive data collection strategy to create a dataset with rich variations in driving scenarios, illumination, and ground truth object TTC distributions. As visualized in Fig.~\ref{fig:data_map}, our collection route spans diverse environments, with colored regions indicating the raw data selected for the final dataset.

The dataset primarily covers highway and urban scenarios, each offering distinct advantages for autonomous driving research. Highway scenes feature high-speed travel and abrupt velocity changes, providing a wide range of object TTCs that are crucial for studying TTC estimation. These scenarios are particularly valuable as they represent life-critical situations for autonomous driving systems. Urban scenes complement this by introducing a greater diversity of traffic participants, including vulnerable road users (VRUs) such as pedestrians and motorcycles, thereby enriching the object category variation in our dataset.
To overcome the limitations of natural driving data and ensure a balanced TTC distribution, we use a proactive driving strategy. Data collectors are instructed to intentionally vary their speed, accelerating and decelerating to generate a wide range of ground truth object TTC values. This approach particularly enhances the density of critical TTC ranges (0-6 seconds), which are typically underrepresented in natural driving scenarios where drivers maintain constant safety distances.

We intentionally capture a comprehensive range of lighting conditions. Data collection occurs across different times (daytime, nighttime, and twilight) and weather conditions (sunny and cloudy). 
To include scenes with extreme illumination transitions to leverage the high dynamic range capabilities of event cameras, we capture scenes involving tunnel traversals and transitions to low-light roads.

Our data collection strategy leads to a dataset that not only addresses the scarcity of collision-imminent scenarios in existing benchmarks but also encompasses a broad spectrum of autonomous driving applications, from highway cruising to complex urban driving scenarios. The dataset's design ensures a robust representation of critical driving situations while maintaining diversity in environmental conditions and traffic scenarios.

\subsection{Annotation}
\begin{figure*}[t]
    \centering
    \includegraphics[width=0.95\textwidth]{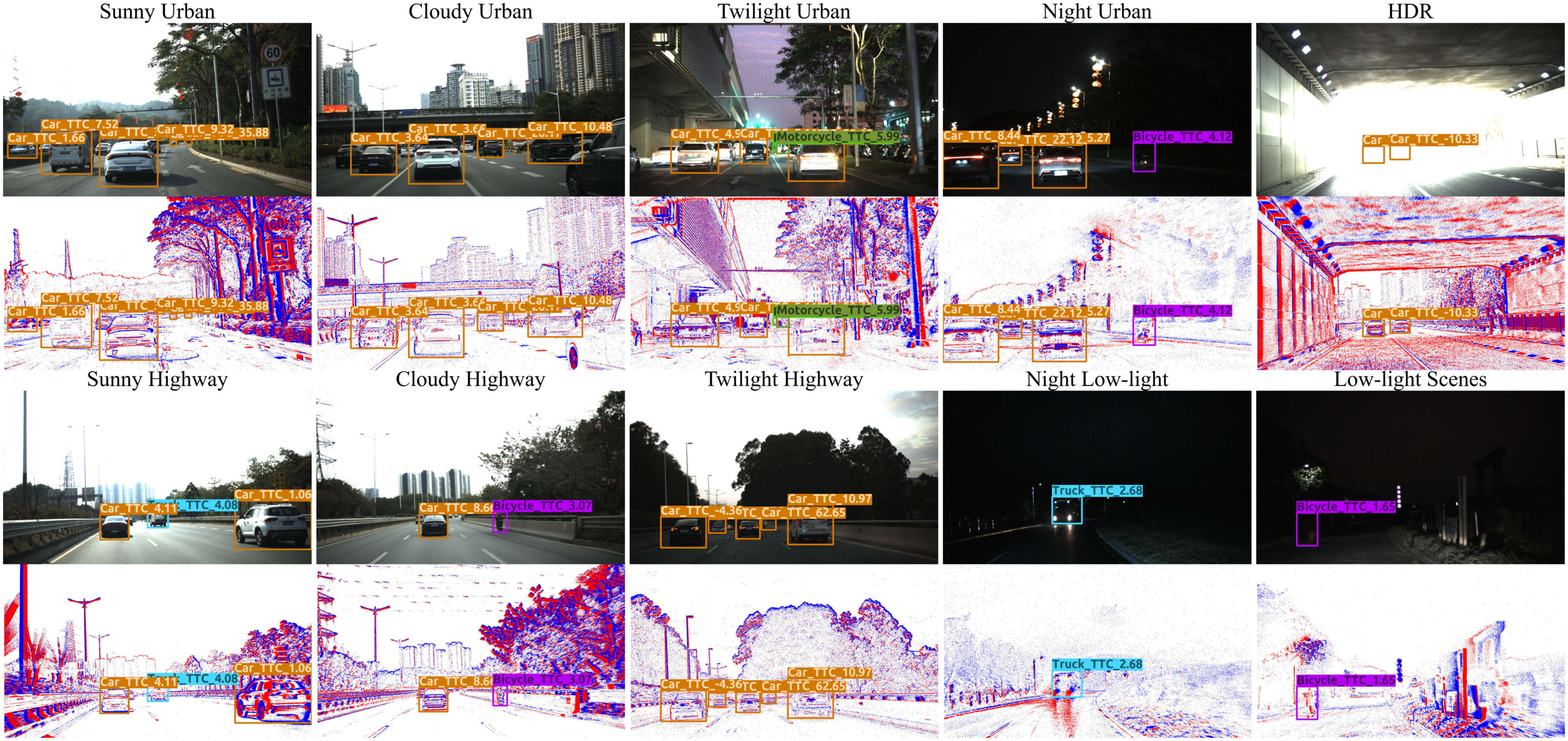}
    \caption{
    Exemplar frames, event data, and object TTC annotations from the \emph{eAP} dataset under varying weather, location, time, and illumination conditions. 
    The 2D bounding boxes show projections of the annotated 3D boxes, along with categories and ground truth TTC values. 
    Event data demonstrates superior quality in over-exposed and low-light scenes, highlighting the benefits of the event modality. Bounding boxes of objects with high visibility are visualized, and a complete visualization of all boxes is included in the supplementary video.
    }
    \label{fig:dataset_example}
\end{figure*}

The data annotation process consists of two phases: a pre-labeling phase and a manual verification phase performed by annotators.

\noindent \textbf{Pre-labeling.} In the pre-labeling phase, we use a 3D object detection model to detect objects from point cloud sequences and link the detected objects into trajectories based on~\cite{weng20203d}. We use BEVFusion~\cite{liu2023bevfusion} as our 3D object detector and train it with 500K frames of our in-house collected data with manual 3D bounding box labels. The detector consists of backbone modules to extract camera and LiDAR features, a view transform module to construct BEV features from camera features, a convolutional neural network (CNN) to fuse the multi-modal features, and a head module for 3D object detection. The prediction head in the detector regresses for each object proposal $p$ its 3D location $(x_p, y_p, z_p)$, 3D size $(l_p, h_p, w_p)$ and yaw angle $\theta_p$ for each frame. 
\jhang{
Objects detected in the vehicle coordinate system are transformed into the world coordinate system using the acquired ego-pose in the world frame. This transformation enables robust tracking in a global reference frame.} 
A tracker based on a 3D Kalman Filter~\cite{weng20203d} maintains a set of trajectories and uses the detections at each timestep to update them. The update step consists of a data association step that matches the detections to the trajectories and a state update step in the Kalman Filter. The state of an object at time $t$ is an 11-dimensional vector $T = (x^t, y^t, z^t, \theta^t, l, w, h, v_x^t, v_y^t, v_z^t, v_{\theta}^t)$ representing its location, orientation, size, velocity, and angular velocity. After processing the whole sequence, we obtain 3D object trajectories with unique object identities and estimated velocities. 

\final{
To ensure high-fidelity ground truth TTC labels, these trajectories undergo offline temporal smoothing to reduce stochastic measurement noise and detection jitter. This refinement improves label stability without incorporating non-causal information or altering the physical dynamics of the observed maneuvers. Crucially, it preserves non-linear behaviors—such as emergency braking and sudden lane changes—that remain unpredictable using naive linear extrapolation.
Furthermore, event-enhanced perception goes beyond simple temporal prediction. When frame-based observations degrade or fail under extreme illumination or occlusions, event cameras provide robust, continuous measurements essential for reliable TTC estimation—something that linear filtering cannot recover from corrupted frame data.}
As shown in Fig.~\ref{fig:lidar_bev}, the 3D bounding boxes in our dataset are accurately labeled in the point cloud. 
Fig.~\ref{fig:dataset_example} further provides sample data from the \emph{eAP} dataset, showing synchronized frames, event streams, and TTC annotations across various conditions.
These visualizations show the 2D projections of the 3D boxes and demonstrate the superior quality of event data in over-exposed and low-light scenes.

\noindent \textbf{Manual checking} 
Annotators are provided with pre-labeled trajectories to verify their quality and perform manual corrections. This process includes correcting inaccuracies in 3D bounding boxes and rectifying ID switches. Annotators also validate the motion profile of each trajectory, ensuring the estimated speed curve is smooth and consistent with the object's motion. The TTC ground truth for object $o$ at time $t$ is obtained by
\begin{equation}
  \tau_o^t = \frac{\min(Z_{o}^{t})}{v^t_{rel}},
\end{equation}
where $\min(Z_{o}^{t})$ is the depth of the nearest point of the object, and $v^t_{rel}$ is the relative speed between it and the ego vehicle in the depth direction. The relative speed vector is obtained from the estimated object velocity in the world coordinate system and the ego vehicle's velocity in the world coordinate system. \rebuttal{This definition of TTC builds upon the geometry proposed in previous classical studies~\cite{nelson1989obstacle}, and also inspires our learning-based geometry-aware framework for object-level TTC estimation, which will be described in the method section.}

\begin{figure}[t]
    \centering
    \includegraphics[width=0.48\textwidth]{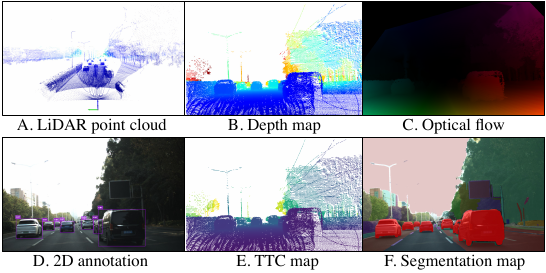}
    \caption{\final{Visualization of additional ground-truth annotation types planned for future releases of the \emph{eAP} dataset, including (A) LiDAR point clouds, (B) depth maps, (C) optical flow, (D) 2D annotations, (E) dense TTC maps, and (F) segmentation maps.}}
    \label{fig:other_anno}
\end{figure}

\noindent \textbf{Future Extensions.} \final{The aforementioned 3D bounding boxes and object-level TTC constitute the core annotations of the current \emph{eAP} release. Beyond this initial focus, our multi-modal sensor suite and high-quality data collection inherently support a wider range of perception tasks. Leveraging the captured dense point clouds and precise ego-localization, we plan to progressively release additional annotation types---such as dense TTC maps, optical flow, and segmentation masks, as visualized in Fig.~\ref{fig:other_anno}. These planned expansions aim to expand \emph{eAP} into a more comprehensive benchmark for the broader research community.}

\begin{figure*}[t]
    \centering
    \includegraphics[width=0.95\textwidth]{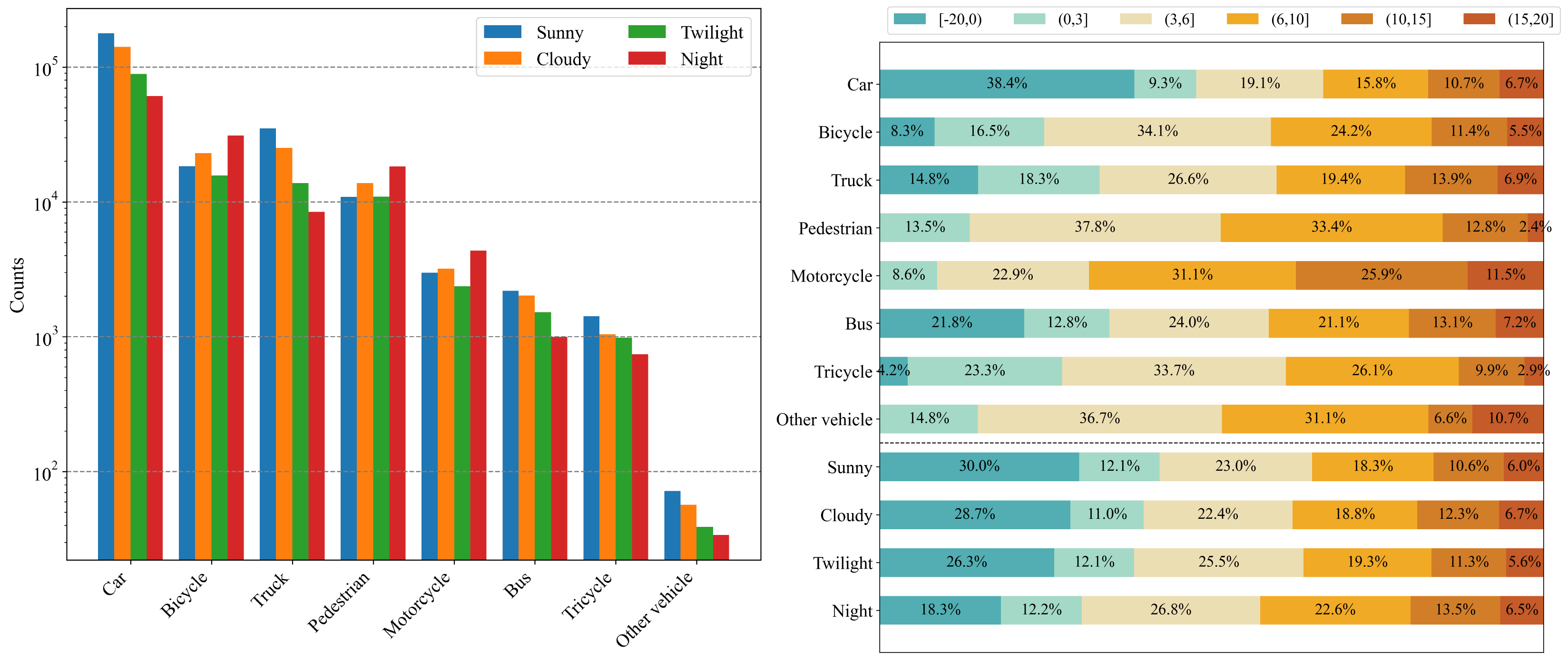}
    \caption{\jhang{Statistics of the \emph{eAP} dataset. 
    Left: distribution of the annotated objects with respect to category and weather. 
    Right: The TTC ground truth distribution shown across the eight object classes and under the four different lighting conditions.}
    }
    \label{fig:dataset_statistics}
\end{figure*}

\begin{table}[t]
    \centering
    \setlength{\tabcolsep}{3pt} %
    \resizebox{0.45\textwidth}{!}{%
    \begin{tabular}{c r c r r} 
        \toprule
        Region & 
        \makecell[c]{Distance [km]} & 
        Illumination & 
        \makecell[c]{Sequences \\ (Train/Test)} & 
        \makecell[c]{Time [min] \\ (Train/Test)} \\
        \midrule
        Highways & 178.44 & 
        \makecell[c]{Sunny \\ Cloudy \\ Twilight} & 
        \makecell[r]{13/3 \\ 10/1\\ 7/2} & 
        \makecell[r]{65/15 \\ 50/5\\ 35/10} \\ 
        \midrule
        Urban & 52.61 km & 
        \makecell[c]{Sunny \\ Cloudy \\ Twilight \\ Night} & 
        \makecell[r]{5/1 \\ 4/1 \\ 1/1 \\ 5/2} & 
        \makecell[r]{25/5 \\ 20/5\\ 5/5 \\ 25/10} \\ 
        \midrule
        Low-light & 
        5.01 km & 
        Night & 
        1/1 & 
        5/5 \\
        \midrule
        \textbf{Total} & 
        \textbf{236.06 km} & 
        -- & 
        \textbf{58} & 
        \textbf{290} \\
        \bottomrule
    \end{tabular}%
    }
    
    \caption{\rebuttal{
    Statistics of the \emph{eAP} dataset, categorized by region and illumination. The table details the total traveled distance [km], along with the train/test split for both sequence counts and duration [min].
    }}
    \label{tab:dataset_stats}
    \vspace{-7ex}
\end{table}

\subsection{Dataset statistics}

As summarized in Tab.~\ref{tab:related_datasets}, \textit{eAP} represents the largest event-based dataset for autonomous driving scenes, comprising 58 sequences and 532k object 3D bounding box and TTC annotations. 
\final{Another key feature of \textit{eAP} is its narrow-baseline RGB-event design. This design reduces the disparity between the two sensors, thereby facilitating robust research on RGB-event fusion for autonomous perception. 
The technical details of our mapping methodology and its resulting sub-5-pixel disparity are presented in the Event-to-RGB mapping paragraph of Sec.~\ref{sec:dataset}.B.}
Fig.~\ref{fig:dataset_statistics} illustrates the distribution of annotated objects across categories, scenes, and ground truth TTC values, highlighting the dataset's coverage of diverse scenarios, including critical cases with small TTC values. 
For benchmarking purposes, \jhang{as shown in Tab.~\ref{tab:dataset_stats}}, the dataset is partitioned into a training set (46 sequences, 138k frames) and a testing set (12 sequences, 36k frames). 
\rebuttal{Our dataset split is performed strictly at the sequence level to ensure no frames from the same sequence appear in both the training and test sets. To ensure the test set provides a fair evaluation of generalization across various environments, we randomly sample representative sequences from each scenario category (e.g., different lighting conditions and regions). This methodology ensures that the test set maintains a balanced distribution of all challenging scenarios.} 
\nicholas{Within the \emph{eAP} test set, we select a fraction of the data captured during travel through tunnels and at nighttime as the HDR subset, in which the illumination is challenging for conventional frame-based SDR cameras. Other data in the test set with normal illumination is denoted as the driving subset.}

\section{Problem formulation: Learning deep representations for event-enhanced perception}
\label{sec:formulation}
Our problem can be formulated in a general Bayes' framework, and each sub-task corresponds to a specification of the general problem. \rebuttal{Given image inputs $\mathcal{I}$ and event inputs $\mathcal{E}$, the 3D perception problem aims to infer the unknown states $a$ as $p(a|\mathcal{I},\mathcal{E}) \propto p(\mathcal{I},\mathcal{E}|a)p(a)$.}
In contrast to the model-based approaches that model the prior term $p(a)$ and likelihood term $p(\mathcal{I},\mathcal{E}|a)$ explicitly, this study takes a discriminative approach~\cite{li2020cascaded} and models the posterior $p(a|\mathcal{I},\mathcal{E})$ with a deep neural network $\mathcal{F}_\Theta$ parametrized by $\Theta$ for inference. Our framework only needs a single forward pass to predict the unknown states, whereas the model-based formulation usually requires careful initialization and iterative optimization to converge. Our core problem is \emph{how to design the function $\mathcal{F}$ to achieve effective performance?} In addition, the network requires a large amount of training pairs $\{(\mathcal{I}_i,\mathcal{E}_i, a_i)\}$ to achieve good generalization performance. Our proposed \emph{eAP} plays an important role to make such a training process possible.

\subsection{The visual 3D vehicle detection task}
Given camera and event images as inputs, the goal of 3D object detection is to detect objects represented as 3D bounding boxes. The multi-modal inputs comprise two parts: (i) an RGB sequence 
$\mathcal{I}^t = \{\mathbf{I}^{j} \mid j = t, t-\Delta t, \ldots, t-K\Delta t\}$, 
and (ii) an event stream $\mathcal{E}^t = \{e_k = (x_k, y_k, t_k, p_k) \mid t-K\Delta t \leq t_k \leq t\}_{k=1}^M$ recording asynchronous pixel-level brightness changes. 
Here, $\Delta t$ denotes the frame interval, $K$ is the temporal window size, and $M$ represents the number of event points within the time window. 
Each event $e_k$ encodes pixel coordinate $(x_k, y_k)$, timestamp $t_k$, and polarity $p_k \in \{0,1\}$. The output is a set of detected objects, where each object has its predicted 3D location $(\hat{x}_o, \hat{y}_o, \hat{z}_o)$, 3D dimension $(\hat{l}_o, \hat{h}_o, \hat{w}_o)$, and yaw angle $\hat{\theta}_o$.

\subsection{The visual object TTC estimation task}
Given sensor observations of an object $o$, this task aims to estimate its TTC $\hat{\tau}_o^t$ at timestamp $t$, defined as the duration until the object reaches the image plane, assuming the current relative motion between the camera and the object remains unchanged. As an instance-level perception task on the FV images, we assume the 2D bounding box inputs $\mathcal{B}_o^t = \{\mathbf{B}_o^{j} \mid j = t, t-\Delta t, \ldots, t-K\Delta t\}$ specifying the object locations on the RGB image sequences, which we use to crop the regions-of-interest $\mathcal{I}_o^t$ and $\mathcal{E}_o^t$. The problem is to learn a parametric function $\mathcal{F}_\Theta$ that maps these heterogeneous data to TTC estimates through $\hat{\tau}_o^t = \mathcal{F}_\Theta(\mathcal{I}_o^t, \mathcal{B}_o^t, \mathcal{E}_o^t)$. The core challenge lies in designing and learning this function for accurate object TTC estimation.

\section{Event-enhanced visual 3D object detection}
\label{sec:method_3dod}

\begin{figure}[t]
    \centering
    \includegraphics[width=0.47\textwidth]{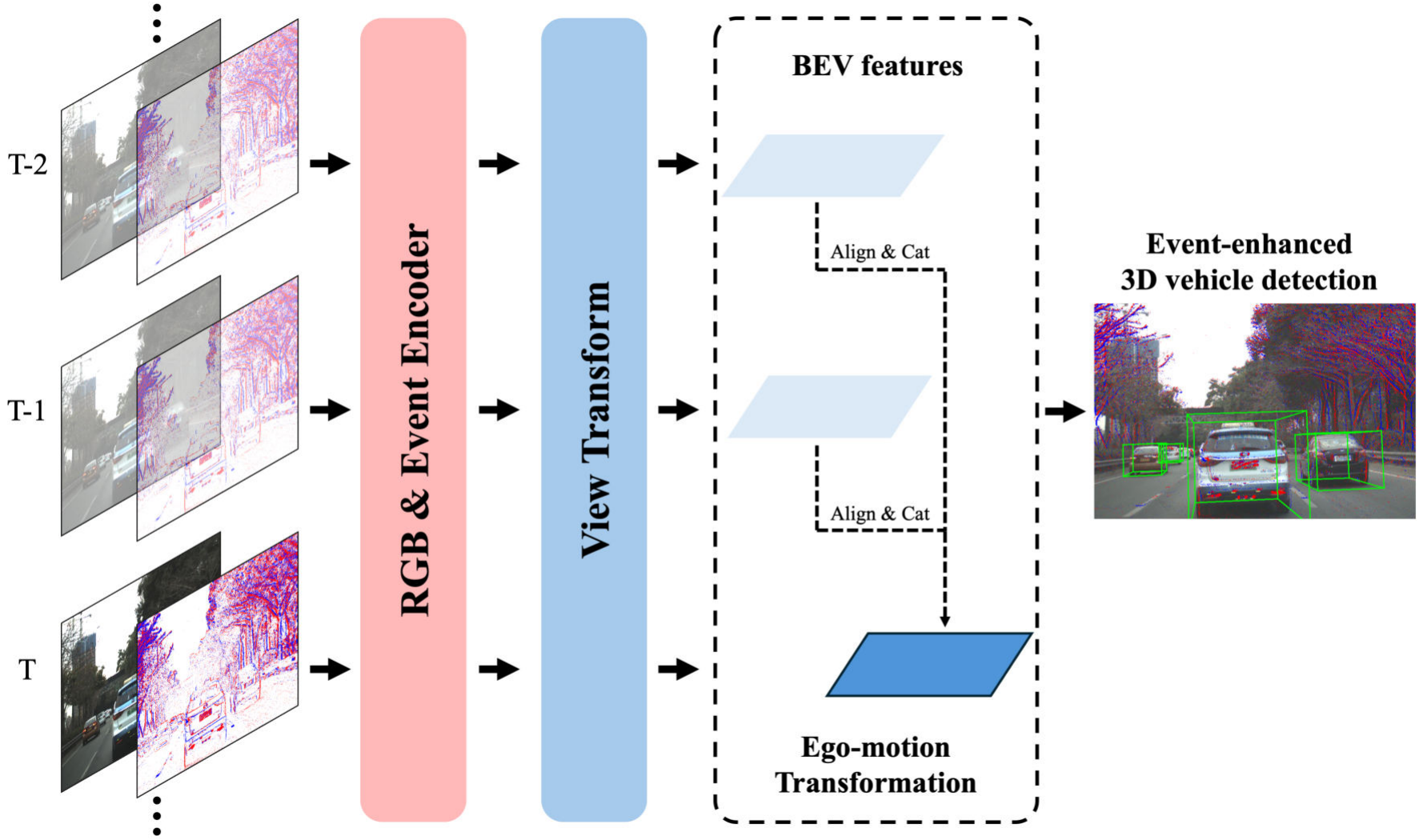}
    \caption{Diagram of the BEV 3D vehicle detection model. The RGB and event encoders extract FV features from the input images and events. The view transform module builds a BEV scene representation based on the FV features. For the event-enhanced model, the FV features extracted from RGB images and events are fused via concatenation. For the temporal model, the BEV features from past frames are warped to the current frame and concatenated with the BEV feature extracted from the current frame. An anchor-based detection head predicts the detected vehicles from the BEV feature.}
    \label{fig:model_architecture_3dod}
    \vspace{-2ex}
\end{figure}

We use a popular visual 3D vehicle detection model~\cite{rukhovich2022imvoxelnet,huang2022bevdet4d} as a baseline and further incorporate information from event cameras for event-enhanced perception.

\noindent\textbf{Event data representation.} 
We employ an event encoding step similar to that in~\cite{zhang2021object}. 
Specifically, events within each time interval $[T_j, T_j + \Delta t]$ are converted into a voxel representation by discretizing the time dimension into $N$ bins.
The events in the $i$th bin are aggregated into a 2D event image $g(x, y, i)$, where each pixel records the polarity of the most recent event at that location.
As a result, we have
\begin{multline}
\label{eq:event}
g(x, y, i) = \frac{p_k\times\delta(t(x, y, i)_{max} - t_k) + 1}{2}\\ 
t(x, y, i)_{max} = max(t_k\times \delta(x-x_k, y-y_k)) \\ 
\forall t_k \in [T_j + (i-1) \Delta t / N, T_j + i\Delta t / N],
\end{multline}
where $\delta$ is the Dirac function.

\noindent\textbf{Model architecture and supervision.}  An overview of our model is shown in Fig.~\ref{fig:model_architecture_3dod}. The detector consists of backbone modules to extract camera and event features, a view transform module to construct BEV features, and a head module for 3D vehicle detection and attribute prediction. 

The backbone modules are implemented with ResNet-50 to obtain FV visual features $\mathcal{F}_{\mathcal{I}}$ and event features $\mathcal{F}_{\mathcal{E}}$. For single-modal models, the FV features are directly passed to the view transform module. For an event-enhanced model that takes multi-modal inputs, we concatenate $\mathcal{F}_{\mathcal{I}}$ with $\mathcal{F}_{\mathcal{E}}$ and use a convolutional network to fuse the two modalities and obtain an event-enhanced FV feature map $\mathcal{F}_{RGBE}$. 

The view-transform module builds a dense voxel grid in the ego vehicle coordinate system. The voxels $\{v_{i,j,k}\}_{i=1,j=1,k=1}^{N_x,N_y,N_z}$ are uniformly sampled along the front (x), left (y), and height (z) directions. The projection of $v_{i,j,k}$ on the RGB image is 
\begin{multline}
p_{i,j,k} = 
\begin{bmatrix}
w*p_{i,j,k}^x\\
w*p_{i,j,k}^y \\
w \\
\end{bmatrix} =
\text{K}_{3 \times 3}v_{i,j,k}
= \text{K}_{3 \times 3}
\begin{bmatrix}
X_{i,j,k}\\
Y_{i,j,k}\\
Z_{i,j,k} \\
\end{bmatrix}
\\
=
\text{K}_{3 \times 3}
\begin{bmatrix}
X_{start} + (i-1)\Delta X\\
Y_{start} + (j-1)\Delta Y \\
Z_{start} + (k-1)\Delta Z \\
\end{bmatrix},
\end{multline}
where $\Delta X=\Delta Y=\Delta Z$ is the voxel resolution. The feature vector for each voxel is then gathered from its projected location on the FV feature map.
\begin{equation}
\mathcal{F}_{i,j,k} = \mathcal{W}(\mathcal{F}_{RGBE}, p_{i,j,k}),
\end{equation} 
where $\mathcal{W}$ denotes the warping operation, and $\mathcal{W}(\mathcal{F}_{RGBE}, p_{i,j,k})$ extracts the feature at $(p_{i,j,k}^x, p_{i,j,k}^y)$ from the corresponding location on the feature maps $\mathcal{F}_{RGBE}$ with bi-linear interpolation. 

The voxel features are down-sampled in the height direction and passed to a CNN-based module for BEV feature extraction. The BEV feature is passed to an anchor-based detection head, which performs anchor classification and regresses the 3D location, 3D size, and yaw angle for each object. During training, the detector is trained by the anchor classification and regression losses $L_{det} = L_{cls} + L_{reg}$.

To enhance the BEV feature map with long-term information, we follow~\cite{huang2022bevdet4d} to build a temporal model. Specifically, for each location $l$ on the BEV feature map at the current frame $\mathcal{F}_{BEV}^t$, we compute its location at a history frame as
\begin{equation}
    l(t-\Delta t) = \text{M}^{t-\Delta T}_{world \rightarrow ego}\text{M}^{t}_{ego \rightarrow world}l(t),
\end{equation}
where $\text{M}$ denotes the transformation between the ego vehicle and the world coordinate system for different timestamps. This location is then used to obtain the historical BEV feature vector similarly by warping $\mathcal{F}_{BEV}^{t-\Delta T}(l(t-\Delta T)) = \mathcal{W}(\mathcal{F}_{BEV}^{t-\Delta T}, l(t-\Delta T))$. For this temporal model, the historical BEV feature maps are concatenated with $\mathcal{F}_{BEV}^{t}$ and passed to the detection head. In this model, the temporal window size $K$ is 2, and $\Delta T$ is 0.1 s. 

\section{Event-Enhanced Visual TTC Estimation}
\label{sec:method_ttc}

We start with a simple baseline, based on which we propose our \emph{Garl-TTC} framework (\textbf{G}eometry-\textbf{A}ware \textbf{R}epresentation \textbf{L}earning for object TTC estimation).

\begin{figure}[t]
    \centering
    \includegraphics[width=0.48\textwidth]{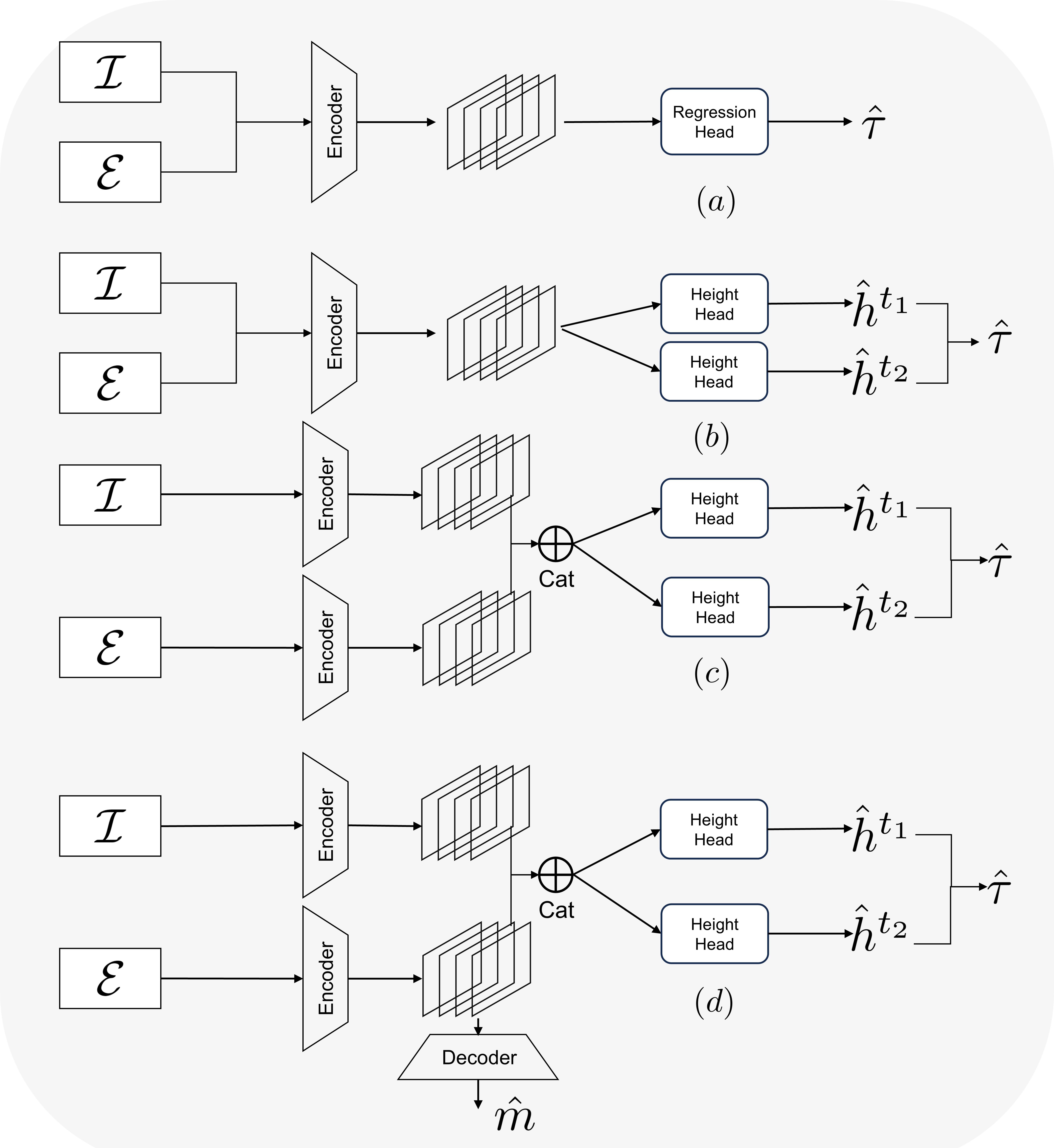}
    \caption{
    Architecture of (a) the baseline model for direct TTC regression, (b) the height-ratio network leveraging TTC geometry, (c) a late-fusion strategy, and (d) an extended version incorporating object foreground estimation during training, supervised by a teacher model.}
    \label{fig:model_architecture}
    \vspace{-1ex}
\end{figure}

\subsection{Baseline}
Our baseline adopts a design similar to previous learning-based approaches~\cite{zhang2021object,chen2020end,gehrig2023recurrent}. 
It comprises three key components: (1) a pre-processing step to transform the event stream into model inputs, (2) an encoder to extract spatial and temporal features from both RGB images and event data, and (3) a regression head for predicting TTC. The event representation approach is the same as that in the 3D vehicle detection task. The temporal window size $K$ is set to 1, i.e., only two frames of RGB images are used as inputs for efficiency.

\noindent\textbf{Model architecture and supervision.}
The baseline model, shown in Fig.~\ref{fig:model_architecture}(a), adopts an early-fusion scheme by concatenating the RGB images and event voxels as inputs. 
It employs a ResNet-50 backbone to extract spatial-temporal features, which are then fed into a fully-connected layer for TTC regression. 
During training, the model is supervised using the ground truth TTC $\tau_o^t$ and optimized with a smooth L1 (SL1) loss function\footnote{
\url{https://docs.pytorch.org/docs/stable/generated/torch.nn.SmoothL1Loss.html}} as
\begin{align}
\label{eq:lossl1}
L_{ttc} & = 
\begin{cases}
\frac{1}{2}(\tau_o^t - \hat{\tau}_o^t)^2 = \\ \frac{1}{2}(\tau_o^t - \mathcal{F}_\Theta(\mathcal{I}_o^t, \mathcal{B}_o^{t}, \mathcal{E}_o^t))^2 ,
&\text{if } |\tau_o^t - \hat{\tau}_o^t| < 1,\\
|\tau_o^t - \hat{\tau}_o^t| - \frac{1}{2}, & \text{else}.
\end{cases}
\end{align}

\subsection{Learning the height-ratio for TTC estimation}
Both the RGB and event data exhibit significant appearance variations caused by differences in object categories and motion patterns, challenging lighting conditions, and sensor noise. 
The straightforward design of function $\mathcal{F}$ in the baseline model is highly non-linear and, as demonstrated in our experiments, proves difficult to learn. 
To address this issue, we propose an alternative function design that leverages the geometric principles of TTC and incorporates intermediate visual representations. 
This approach enables the model to learn more effectively from the sensor inputs.

\begin{figure}[t]
    \centering
    \includegraphics[width=0.5\textwidth]{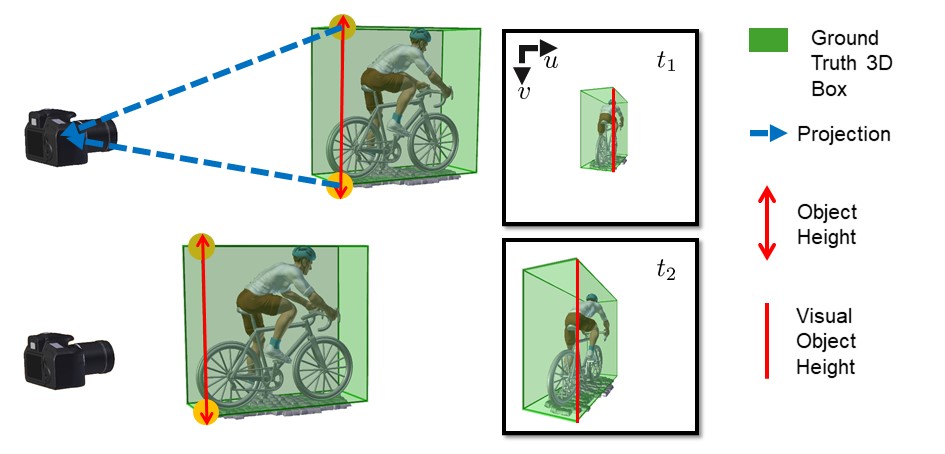}
    \caption{Diagram illustrating the change in visual object height from time $t_1$ to $t_2$ as the object moves closer to the camera.}
    \label{fig:ttc_diagram}
    \vspace{-1ex}
\end{figure}

To formalize this, we begin by considering an object $o$ at time $t_1$ represented as a point set in the camera coordinate system: $o^{t_1} = \{(X_{i}^{t_1}, Y_{i}^{t_1}, Z_{i}^{t_1}) | \ i=1,2,...,P\}$.
We introduce two key assumptions about the object motion and scene geometry in autonomous driving scenarios.

\noindent\textbf{Rigid translation motion}: 
Over a small time interval $\Delta t$, the object's 3D translation with respect to the camera updates the coordinates of its points as: $o^{t_2} = \{ (X_i^{t_1}+\delta_x, Y_i^{t_1}+\delta_y, Z_i^{t_1}+\delta_z) \}_{i=1}^P$, where $\{\delta_x, \delta_y, \delta_z\}$ denotes the translational increment.
Let $i^* = \arg\min_i (\{Z_i^{t_1} | \ i=1,2,...,P\})$ denote the index of the nearest point (smallest $Z$-coordinate) at time $t_1$.
Since the rigid translation motion assumption enforces a uniform translation $\delta_z$ for all points, the relative depth ordering remains unchanged, namely for any $j \neq i^*$ we have
$Z_j^{t_2} = Z_j^{t_1} + \delta_z \geq Z_{i^*}^{t_1} + \delta_z = Z_{i^*}^{t_2}$.
Thus, $i^*$ remains the nearest point at $t_2$.
The object-level TTC can, therefore, be approximated by the instantaneous TTC of the nearest point:
\begin{equation}
\tau_o^t = \frac{\Delta_{t} \cdot Z_{\min}^{t_1}}{Z_{\min}^{t_1} - Z_{\min}^{t_2}},
\label{eq:ttc}
\end{equation}
where $Z_{\min}^{t}$ denotes the smallest depth in the set $\{Z_i^{t}\}$.

\noindent\textbf{Vertical height assumption}: In practical autonomous driving scenes, most objects move on the ground plane. 
We assume each object has an object height represented by a vector parallel to the camera's y-axis and has norm $H$. 
The observed object height (in pixels) on the image plane is
\begin{equation}
h^{t_1} = \frac{f_y \cdot H}{Z_{\min}^{t_1}}, \label{eq:height_prior}
\end{equation}
where $f_y$ denotes the focal length. 
By combining Eq.~\ref{eq:ttc} and Eq.~\ref{eq:height_prior}, we have
\begin{equation}
\tau_o^t = \frac{\Delta_{t}}{1 - \frac{Z_{\min}^{t_2}}{Z_{\min}^{t_1}}} = 
\frac{\Delta_{t}}{1 - \frac{h^{t_1}}{h^{t_2}}}.
\end{equation}
Thus, the object TTC is strongly correlated with the ratio of visual heights as shown in Fig.~\ref{fig:ttc_diagram}. 
Since the visual height can be easier to learn from the rich semantic spatial-temporal features extracted by the backbone module, we propose to design a height-ratio network as 
\begin{equation}
\hat{\tau}_o^t = \mathcal{H}_{\Theta}(\mathcal{I}_o^t, \mathcal{B}_o^{t}, \mathcal{E}_o^t) = \frac{\Delta_{t}}{1 - \frac{\hat{h}^{t_1}_{\Theta}(\mathcal{I}_o^t, \mathcal{B}_o^{t}, \mathcal{E}_o^t)}{\hat{h}^{t_2}_{\Theta}(\mathcal{I}_o^t, \mathcal{B}_o^{t}, \mathcal{E}_o^t)}},    
\end{equation}
where $\hat{h}^{t_1}_{\Theta}$ is the regressed visual object height at $t_1$. 
The design of this height-ratio network is shown in Fig.~\ref{fig:model_architecture}(b). This study uses $K=1$, and the model infers the visual object heights at time $t_1 = t - \Delta t$ ($t = t_2$ in our case) to obtain the estimated TTC $\hat{\tau}_o^t$.
During training, we penalize the predicted height-ratios with the ground truth as
\begin{equation}
    L_{hr} = \log(\frac{\hat{h}^{t_1}_{\Theta}(\mathcal{I}_o^t, \mathcal{B}_o^{t}, \mathcal{E}_o^t)}{\hat{h}^{t_2}_{\Theta}(\mathcal{I}_o^t, \mathcal{B}_o^{t}, \mathcal{E}_o^t)}
    ) - \log(\frac{h^{t_1}}{h^{t_2}}).
\end{equation}

\begin{figure*}[t!]
    \centering
    \includegraphics[width=0.9\linewidth]{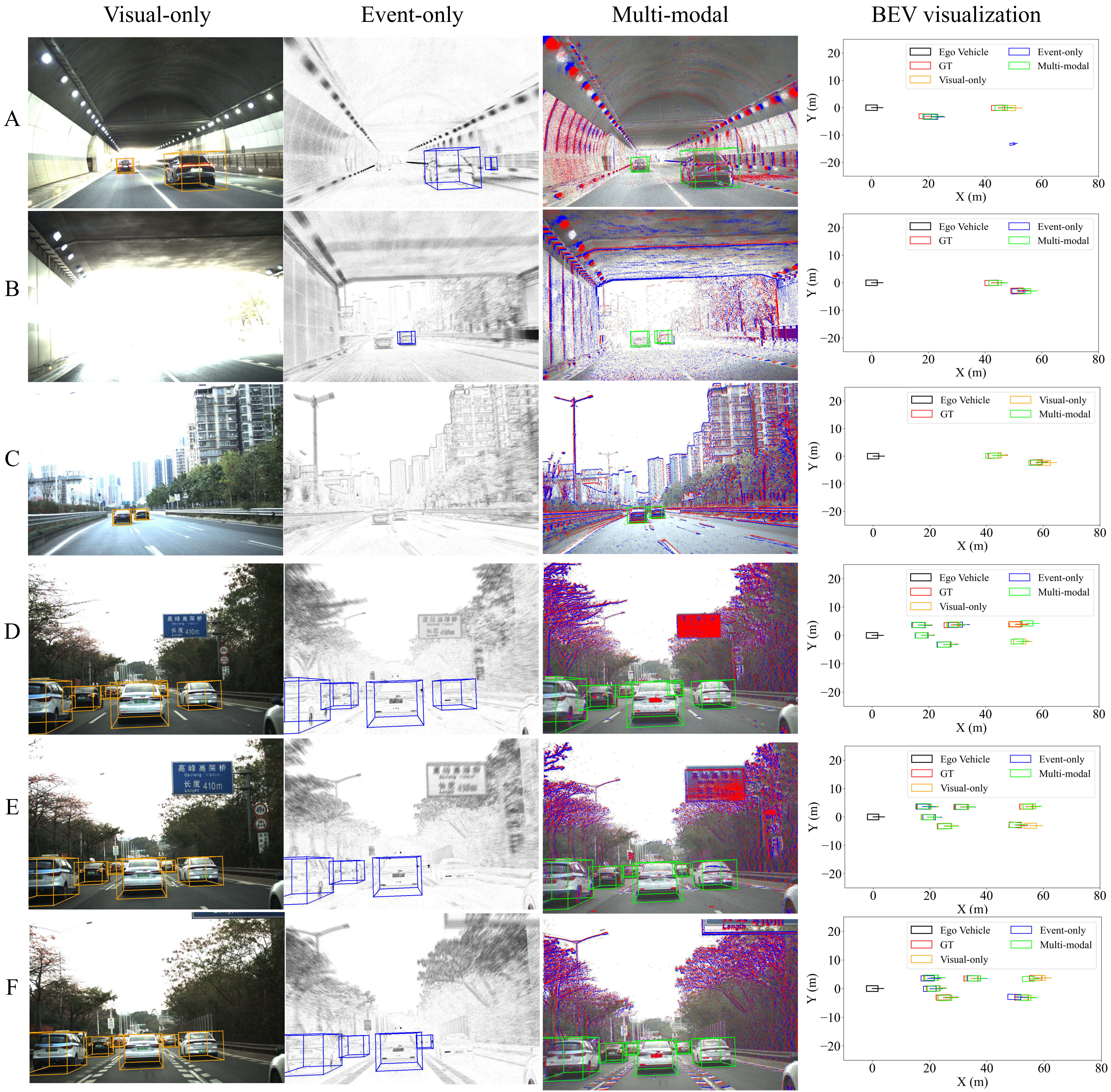}
    \caption{
    \rebuttal{Qualitative visualization of 3D vehicle detection performance on the \emph{eAP} testing set. The figure presents two representative sequences (Frames A-C and D-F). 
    The first three columns compare the detection results generated from different inputs: 
    \textbf{(a)} Visual-only, \textbf{(b)} Event-only, and \textbf{(c)} Our proposed event-enhanced multi-model. 
    The fourth column \textbf{(d)} provides a Birds-Eye-View (BEV) visualization, 
    comparing the predicted boxes with the ground truth. 
    Best viewed in color and zoomed in for more details.}}
    \label{fig:bev_compare}
    \vspace{-1ex}
\end{figure*}

\subsection{Learning representation to super-resolve the object foreground}
The object TTC is highly correlated with changes in visual object heights, which, in turn, are strongly influenced by the object's silhouette. 
Intuitively, if a model learns a latent representation capable of generating high-resolution object boundaries, such a representation would be highly effective for accurate visual height estimation, ultimately leading to precise TTC estimation.
To this end, we propose introducing an additional supervision mechanism to encourage the model to perform high-resolution foreground segmentation. 
Specifically, we augment the model with a decoder, as illustrated in Fig.~\ref{fig:model_architecture}(d). 
The input resolution is set to $128 \times 128$ pixels, while the model is tasked with generating an object foreground mask $\hat{m}$ at a higher resolution of $256 \times 256$ pixels.
For supervision, we leverage SAM~\cite{kirillov2023segment} as a teacher model and employ a knowledge distillation loss formulated as a cross-entropy loss: $L_{KD} = \text{CE}(\hat{m}, m_{SAM})$, where $m_{SAM}$ denotes the object foreground predicted by SAM. Notably, the decoder and the associated supervision are utilized exclusively during training, ensuring that no additional computational costs are incurred during inference.

\section{Experiments}
\label{sec:experiment}

\vspace{1ex}
\begin{table*}[t!]
\centering
\resizebox{0.9\linewidth}{!}{%
\normalfont
\begin{tabular}{lllllllllll}
\toprule
  \multicolumn{1}{l}{\multirow{2}{*}{Methods}} &
  \multicolumn{1}{c}{\multirow{2}{*}{Modality}} &
  \multicolumn{1}{c}{\multirow{2}{*}{\makecell[c]{Frame \\Number}}} &
  \multicolumn{4}{c}{Driving} &
  \multicolumn{4}{c}{HDR} \\
\cmidrule(l{1mm}r{1mm}){4-7}
\cmidrule(l{1mm}r{1mm}){8-11}
\multicolumn{1}{c}{} &
\multicolumn{1}{c}{} &
\multicolumn{1}{c}{} &
\multicolumn{1}{l}{AP$~\uparrow$} &
\multicolumn{1}{l}{ATE$~\downarrow$} &
\multicolumn{1}{l}{ASE$~\downarrow$} &
\multicolumn{1}{l}{AOE$~\downarrow$} &
\multicolumn{1}{l}{AP$~\uparrow$} &
\multicolumn{1}{l}{ATE$~\downarrow$} &
\multicolumn{1}{l}{ASE$~\downarrow$} &
\multicolumn{1}{l}{AOE$~\downarrow$}  \\ \midrule

Visual-only &
  \multicolumn{1}{c}{V} &
  \multicolumn{1}{c}{1} &
  0.510 &
  0.497 &
  \uline{0.075} & 
  \uline{0.021} &
  0.403 &  
  0.584 & 
  \textbf{0.018} &
  0.087
\\ 

Event-only &
  \multicolumn{1}{c}{E} &
  \multicolumn{1}{c}{1} &
  0.200 &
  0.794 &
  0.096 & 
  0.023 &
  0.138 &
  0.872 &
  0.118 &
  0.031
  \\ 

Fusion &
  \multicolumn{1}{c}{V+E} &
  \multicolumn{1}{c}{1} &
  \uline{0.515} &
  \uline{0.482} &
  0.076 & 
  \uline{0.021} &
  \uline{0.460} &
  \uline{0.503} &
  \uline{0.082} &
  \uline{0.020}
  \\ 

Fusion-temporal &
  \multicolumn{1}{c}{V+E} &
  \multicolumn{1}{c}{3} &
  \textbf{0.531} &
  \textbf{0.400} &
  \textbf{0.072} & 
  \textbf{0.018} &
  \textbf{0.558} &
  \textbf{0.363} &
  0.085 &
  \textbf{0.015}
  \\
  
\midrule

\rebuttal{CenterNet3D Visual-only} &
\multicolumn{1}{c}{\rebuttal{V}} &
\multicolumn{1}{c}{\rebuttal{1}} &
\rebuttal{0.423} &
\rebuttal{\uline{0.495}} &
\rebuttal{0.099} &
\rebuttal{0.027} &
\rebuttal{0.389} &
\rebuttal{\uline{0.511}} &
\rebuttal{\textbf{0.099}} &
\rebuttal{0.023}
\\

\rebuttal{CenterNet3D Event-only} &
\multicolumn{1}{c}{\rebuttal{E}} &
\multicolumn{1}{c}{\rebuttal{1}} &
\rebuttal{0.208} &
\rebuttal{0.726} &
\rebuttal{0.119} &
\rebuttal{\textbf{0.023}} &
\rebuttal{0.149} &
\rebuttal{0.835} &
\rebuttal{0.139} &
\rebuttal{0.023} 
\\

\rebuttal{CenterNet3D Fusion} &
\multicolumn{1}{c}{\rebuttal{V+E}} &
\multicolumn{1}{c}{\rebuttal{1}} &
\rebuttal{\uline{0.480}} &
\rebuttal{0.508} &
\rebuttal{\uline{0.095}} &
\rebuttal{\uline{0.024}} &
\rebuttal{\uline{0.428}} &
\rebuttal{0.563} &
\rebuttal{\uline{0.102}} &
\rebuttal{\uline{0.015}}
\\

\rebuttal{CenterNet3D Fusion-temporal} &
\multicolumn{1}{c}{\rebuttal{V+E}} &
\multicolumn{1}{c}{\rebuttal{3}} &
\rebuttal{\textbf{0.493}} &
\rebuttal{\textbf{0.471}} &
\rebuttal{\textbf{0.092}} &
\rebuttal{\textbf{0.023}} &
\rebuttal{\textbf{0.465}} &
\rebuttal{\textbf{0.449}} &
\rebuttal{0.110} &
\rebuttal{\textbf{0.013}}

   \\ \bottomrule
\end{tabular}
}
\caption{\rebuttal{Quantitative evaluation of 3D vehicle detection performance on the \emph{eAP} testing set for the Car category using our anchor-based model and another anchor-free model, CenterNet3D~\cite{yin2021center}. V: Visual (RGB), E: Event, V+E: Visual (RGB) + Event. 
Best and second best results are \textbf{highlighted} and \uline{underlined}, respectively. 
Models trained using different input modalities and numbers of input frames are compared in different driving scenes. Driving indicates normal scenes while HDR indicates driving scenes with challenging illumination.}}
\label{tab:ablation_3dod}
\vspace{-1ex}
\end{table*}

We begin by introducing the benchmarks and metrics used for quantitative evaluation. We then present qualitative and quantitative results of the deep models trained using our \emph{eAP} dataset. We further conduct a comprehensive comparison of the event-enhanced TTC estimation model with state-of-the-art methods. Finally, we conduct an extensive ablation study to validate the effectiveness of our proposed deep representation learning framework. 

\subsection{Benchmarks and metrics}
\noindent\textbf{Evaluation for visual 3D vehicle detection.} Our \emph{eAP} dataset is used for the first quantitative evaluation of the effectiveness of event-enhanced visual 3D vehicle detection. We follow the metrics of nuScenes~\cite{caesar2020nuscenes} and adapt the official toolkit\footnote{https://github.com/nutonomy/nuscenes-devkit} for quantitative evaluation. Specifically, the standard \emph{Average precision} (AP) metric~\cite{geiger2012we, everingham2010pascal} is used to evaluate the detection performance, where a match is defined by thresholding the 2D center distance on the ground plane. We also use the \emph{True Positive metrics} in~\cite{caesar2020nuscenes} such as \emph{Average Orientation Error} (AOE), \emph{Average Translation Error} (ATE), and \emph{Average Scale Error} (ASE) to evaluate the orientation, location, and dimension estimation accuracy of the predictions that are matched with a ground truth box. ATE is the Euclidean center distance in 2D (units in \text{m}). ASE is computed from the 3D intersection over union (IOU) after aligning orientation and translation (1 - \text{IoU}). AOE is the smallest yaw angle difference between prediction and ground truth (radians).

\noindent\textbf{Evaluation for visual object TTC estimation.} To evaluate the object TTC estimation accuracy, we use our \textit{eAP} dataset as well as the previously established FCWD dataset~\cite{li2024strttc} and \rebuttal{EvTTC~\cite{sun2024evttc} dataset}. Following previous studies~\cite{yang2020upgrading,TSTTC,badki2021binary}, we employ the Motion-in-Depth error (MiD) as our primary evaluation metric, defined as 
\begin{equation}
\text{MiD} = |\log(\hat{\eta}) - \log(\eta)| \times 10^{4},
\end{equation}
where $\hat{\eta}$ and $\eta$ represent the predicted and ground-truth object motion-in-depth, respectively.
The object motion-in-depth is related to the TTC by $\eta = 1- \frac{\Delta t}{\tau}$. In addition, we also use the relative TTC error as our metric, defined as
$
\text{RTE} = \vert \frac{\tau - \hat{\tau}}{\tau} \vert \times 100\%
$, where $\tau$ denotes the ground-truth TTC value and $\hat{\tau}$ the estimated TTC value.%
We present evaluation results across four distinct ranges of TTC ground truths:
\emph{crucial} ($\tau \in (0,3]$), \emph{small} ($\tau \in (3,6]$), \emph{large} ($\tau \in (6,10]$), and \emph{negative} ($\tau \in [-10,0)$). 
The overall performance is computed as a weighted sum of the performance across each TTC range. To emphasize scenarios with small TTC values, which may indicate a higher risk of collision, we assign weights of 0.5, 0.3, 0.1, and 0.1 to the \emph{crucial}, \emph{small}, \emph{large}, and \emph{negative} ranges, respectively.

\vspace{1ex}
\begin{table*}[t]
\centering
\resizebox{\linewidth}{!}{%
\normalfont
\begin{tabular}{lccrrrrrrrr}
\toprule 
Methods  
& \rebuttal{Type}
& \rebuttal{Modality} 
&MiD$_{\text{c}}\downarrow$ (FR$\downarrow$)
&MiD$_{\text{s}}\downarrow$ (FR$\downarrow$)
&MiD$_{\text{l}}\downarrow$ (FR$\downarrow$)
&MiD$_{\text{n}}\downarrow$ (FR$\downarrow$)
&RTE$_{\text{c}}\downarrow$
&RTE$_{\text{s}}\downarrow$
&RTE$_{\text{l}}\downarrow$
&RTE$_{\text{n}}\downarrow$
\\ \midrule 

FAITH~\cite{dinaux2021faith} &
\rebuttal{M} &
\rebuttal{E} &
606.8 (1.2\%)& 
490.8 (0.9\%)& 
319.0 (0.9\%)& 
376.4 (0.5\%)& 
282.7& 
311.8& 
515.7&
428.9
\\ %

ETTCM$_{\text{scaling-3}}$~\cite{nunes2023time} &
\rebuttal{M} &
\rebuttal{E} &
402.2 (0.0\%)& 
\final{\uline{279.5 (0.0\%)}}& 
\final{\uline{263.9 (0.1\%)}}& 
\final{\uline{207.5 (0.0\%)}}& 
414.5& 
181.6& 
288.2&
132.9
\\ 

ETTCM$_{\text{six-dof-3}}$~\cite{nunes2023time} &
\rebuttal{M} &
\rebuttal{E} &
\final{\uline{226.1 (0.0\%)}}& 
326.2 (0.0\%)& 
321.6 (0.0\%)& 
223.8 (0.0\%)& 
222.6& 
17703.4& 
784.8&
660.1
\\ %

CMax~\cite{gallego2018unifying} &
\rebuttal{M} &
\rebuttal{E} &
632.8 (6.7\%) & 
1583.6 (5.5\%)& 
1528.0 (9.6\%)& 
1187.0 (7.9\%)& 
48.2  & 
107.3 & 
\final{\uline{132.4}} & 
235.7 
\\ 

STRTTC~\cite{li2024strttc} &
\rebuttal{M} &
\rebuttal{E} &
237.2 (0.0\%) & 
532.9 (2.7\%) & 
954.3 (21.6\%) & 
348.9 (5.3\%) & 
\final{\uline{37.3}}  & 
\final{\uline{54.1}}  & 
209.4 & 
\final{\uline{48.3}}  
\\ \midrule %

\textbf{\emph{Garl-TTC} (Ours)} & 
\rebuttal{L} &
\rebuttal{E+V} &
\textbf{53.1 (0.0\%)} & 
\textbf{37.6 (0.0\%)} & 
\textbf{40.6 (0.0\%)} & 
\textbf{31.3 (0.0\%)} & 
\textbf{16.6} & 
\textbf{20.0}  & 
\textbf{34.1} & 
\textbf{28.2}
\\ \bottomrule %
\end{tabular}
}

\caption{
Comparison of Motion-in-depth Error (MiD) for different ranges of TTC ground truth \rebuttal{on the \emph{eAP} testing set}. \rebuttal{Failure Ratio (FR)} indicates the proportion of invalid TTC predictions of previous approaches that fall out of the evaluation range. \rebuttal{E: Event, V+E: Visual (RGB) + Event, M: Model-based, L: Learning-based.} 
Best and second best results are \textbf{highlighted} and \uline{underlined}, respectively.} %
\label{tab:method_benchmark}

\end{table*}

\begin{figure*}[h]
    \centering
    \includegraphics[width=0.85\linewidth]{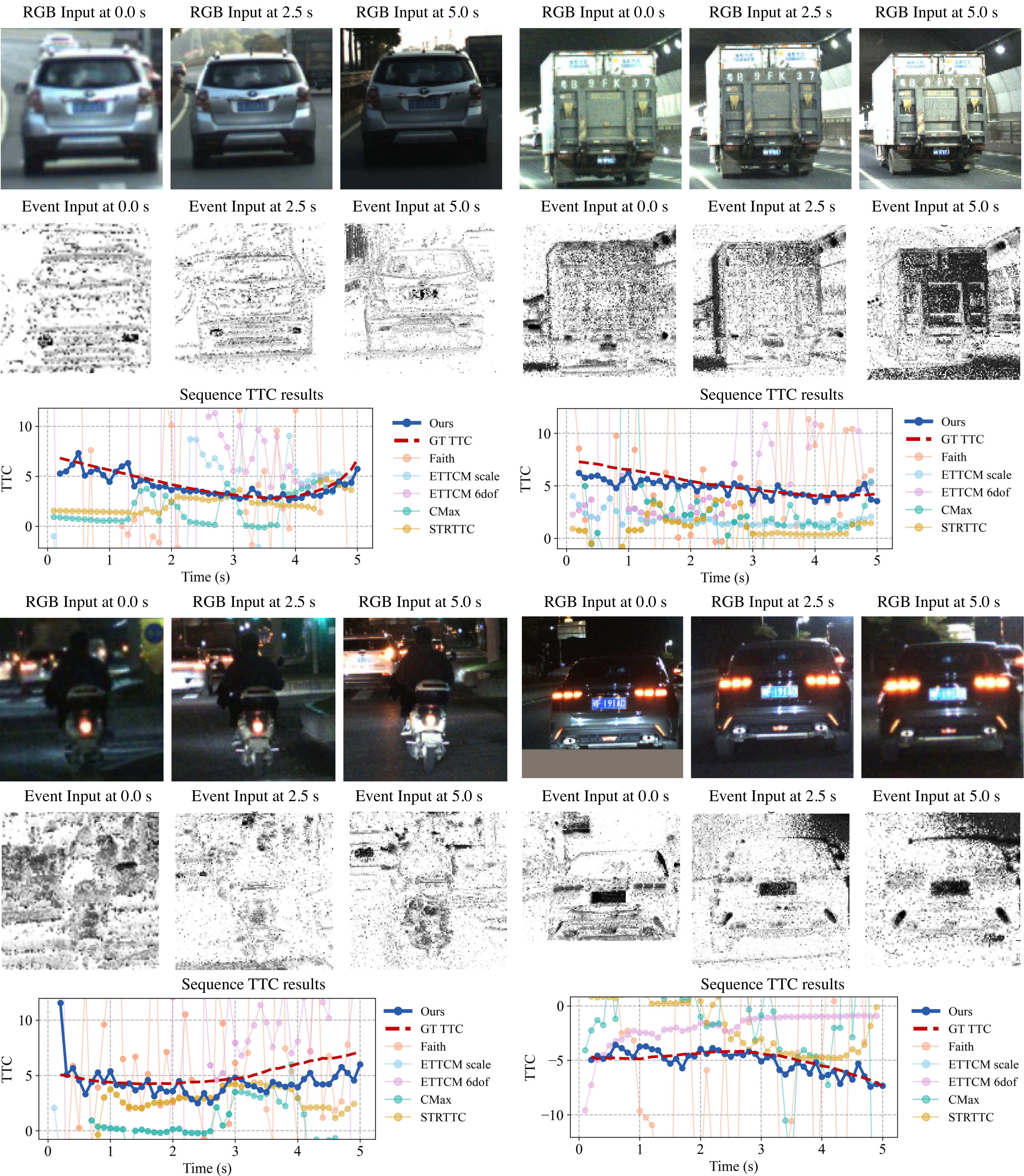}
    \caption{\rebuttal{TTC estimation results of our approach for different object trajectories on the \emph{eAP} dataset. Input frames and event data are shown for several timestamps of the trajectory. 
    Our predicted and ground-truth TTC curves are shown in blue and red, respectively. Best viewed in color.}}
    \label{fig:qualitative_results}

\end{figure*}
\vspace{1ex}
\begin{table*}[t]
\centering
\setlength{\tabcolsep}{3pt}
\footnotesize
\resizebox{0.92\linewidth}{!}{%
\normalfont
\setlength{\tabcolsep}{3pt}
\begin{tabular}{lccrrrrrrrr} 
\toprule
\multicolumn{1}{l}{\multirow{2}{*}{Method}} &
  \multicolumn{1}{c}{\multirow{2}{*}{\final{Type}}} &
  \multicolumn{1}{c}{\multirow{2}{*}{\final{Modality}}} &
  \multicolumn{2}{c}{CCRs2-medium} &
  \multicolumn{2}{c}{CCRs2-high} &  
  \multicolumn{2}{c}{CCRm-medium} &
  \multicolumn{2}{c}{Average} \\  %
\cmidrule(l{1mm}r{1mm}){4-5}
\cmidrule(l{1mm}r{1mm}){6-7}
\cmidrule(l{1mm}r{1mm}){8-9}
\cmidrule(l{1mm}r{1mm}){10-11} %
\multicolumn{1}{c}{} &
  \multicolumn{1}{c}{} &
  \multicolumn{1}{c}{} &
  \multicolumn{1}{c}{RTE~(\%)$\downarrow$} &
  \multicolumn{1}{c}{Runtime~(ms)$\downarrow$} &
  \multicolumn{1}{c}{RTE~(\%)$\downarrow$} &
  \multicolumn{1}{c}{Runtime~(ms)$\downarrow$} &
  \multicolumn{1}{c}{RTE~(\%)$\downarrow$} &
  \multicolumn{1}{c}{Runtime~(ms)$\downarrow$} &
  \multicolumn{1}{c}{RTE~(\%)$\downarrow$} &  %
  \multicolumn{1}{c}{Runtime~(ms)$\downarrow$}  %
   \\ \midrule

FAITH~\cite{dinaux2021faith}  & \final{M} & \final{E}
& 47.45   & 191 
& 49.62 & 165 
& 56.26  &  110
& 51.11 & 155 %
\\

ETTCM~\cite{nunes2023time} & \final{M} & \final{E}
& 52.84 & 98
& 49.46 & 127 
& 58.00  & 121
& 53.43 & 115 %
\\

CMax~\cite{gallego2018unifying} & \final{M} & \final{E}                                     
& \uline{11.20}  &3449
& \uline{11.10} & 3826 
& 14.73 & 3657
& 12.34 & 3644 %
\\

STRTTC~\cite{li2024strttc} & \final{M} & \final{E}   
& 11.83 & \uline{24}
& 13.68 & \uline{28}      
& \textbf{10.36} & \uline{23}
& \uline{11.96} & \uline{25}  %
\\

\midrule

\textbf{\emph{Garl-TTC} (Ours)} & \final{L} & \final{E+V}
 & \textbf{8.31}  & \textbf{13}
 & \textbf{10.56} & \textbf{13}
 & \uline{12.93} & \textbf{12}
 & \textbf{10.60} & \textbf{13} %
\\

\bottomrule

\end{tabular}
}
\caption{\final{\rebuttal{Performance of multiple TTC estimation algorithms on the EvTTC benchmark~\cite{sun2024evttc}. 
\final{E: Event, V+E: Visual (RGB) + Event, M: Model-based, L: Learning-based.}
Best and second best results are \textbf{highlighted} and \uline{underlined}, respectively.}}}
\label{tab:evttcbenchmark_with_avg} %
\end{table*}

\subsection{Event-enhanced BEV 3D vehicle detection}
\subsubsection{Qualitative results}
\rebuttal{Fig.~\ref{fig:bev_compare} shows the 3D vehicle detection results of different models that take RGB images, events, and both as inputs, respectively.} Note that in Fig.~\ref{fig:bev_compare}(B) with challenging illumination, the model using only RGB images as inputs (first row) fails to detect any vehicles. The model using only events (second row) performs better, and the event-enhanced visual model (third row) obtains the best detection recall. This result marks the first successful utilization of events to help improve visual 3D vehicle detection in autonomous driving scenes. 

However, event-based detection performance degrades when the relative speed between the ego-vehicle and leading vehicles is low. This results in a sparse event stream for the object, making detection challenging for event-only models (as shown in Fig.~\ref{fig:bev_compare}(A, E, F)). Thanks to its effective 
fusion of event and visual features, the event-enhanced model still performs robustly in such cases and scenes like Fig.~\ref{fig:bev_compare}(C) where the event-only model fails.

\subsubsection{Quantitative results} Tab.~\ref{tab:ablation_3dod} verifies the above qualitative observations quantitatively. For both normal driving scenes and those with challenging illumination (HDR), the event-only model has worse overall performance than the visual-only model. The event-enhanced model outperforms both the visual-only and event-only models in terms of detection (AP) and location (ATE). Notably, with temporal information aggregation, the temporal model that takes three frames as inputs further improves the overall 3D vehicle detection performance.

\nicholas{To verify that our findings are not restricted to a specific 3D object detector design, we implement another famous 3D object detector, CenterNet3D~\cite{yin2021center}. CenterNet3D uses an anchor-free architecture design for object detection, which is drastically different from our anchor-based approach during training and inference.
We implement different variants of CenterNet3D: 
(a) a visual-only baseline that takes only RGB input, 
(b) an event-only baseline that takes only event input, 
(c) a single-frame fusion model (RGB-E) that uses both visual and event data, and 
(d) a temporal variant based on (c) that uses three frames of inputs.
Despite the large differences in detection architecture, we arrive at similar conclusions. 
The single-frame fusion model (c) achieves better detection performance than 
both the visual-only (a) and event-only (b) baselines. 
Furthermore, aligning with our previous discussion, the event-only model (b) 
underperforms compared to the visual-only model (a). 
The temporal version (d) also outperforms its single-frame counterpart (c). 
These results validate the effectiveness of incorporating event inputs 
across different object detection model architectures.}

\subsection{Event-enhanced visual object TTC estimation}
\subsubsection{Quantitative comparison with previous studies}
Tab.~\ref{tab:method_benchmark} presents the quantitative evaluation of TTC estimation performance for our approach and prior studies \rebuttal{on the \emph{eAP} testing set}. 
Compared to state-of-the-art optimization-based methods~\cite{gallego2018unifying, li2024strttc}, our approach achieves significantly lower MiD and relative TTC errors across all ground truth ranges, highlighting the limitations of optimization-based methods in complex driving scenarios and underscoring the effectiveness of our data-driven deep representation learning. 
Notably, our approach excels in safety-critical scenarios with small TTC values, out-performing existing methods by a substantial margin and demonstrating its potential for autonomous safety applications. 
\rebuttal{Qualitative results in Fig.~\ref{fig:qualitative_results} further illustrate the robustness of our approach, which reliably estimates TTC under challenging conditions such as over-exposure and poor lighting, where prior methods fail.}

\nicholas{A similar quantitative comparison on the EvTTC benchmark~\cite{sun2024evttc} is summarized in Tab.~\ref{tab:evttcbenchmark_with_avg}.}
\rebuttal{For this evaluation, we select three challenging real-world scenarios from the EvTTC benchmark~\cite{sun2024evttc} that feature real vehicles with high relative speeds. We compare our method against several state-of-the-art (SOTA) 
baselines on these challenging scenarios. The non-least-squares nature of CMax~\cite{gallego2018unifying} results in prohibitive computation complexity (avg. 3,644~ms), rendering it unsuitable for real-time applications. ETTCM~\cite{nunes2023time} and FAITH~\cite{dinaux2021faith} also exhibit significantly higher average RTEs (53.43\% and 51.11\%, respectively) in these dynamic environments. STRTTC~\cite{li2024strttc} delivers commendable performance, achieving the best accuracy (10.36\%) on the CCRm-medium scene. 
However, contrasting with its performance on our \emph{eAP} dataset, this success is likely due to the specific characteristics of the EvTTC benchmark. The EvTTC benchmark consists primarily of specific car-following scenarios on straight roads that align well with STRTTC's simplified kinematic assumptions.}

\rebuttal{In sharp contrast, our Garl-TTC, which was trained on our \emph{eAP} dataset and evaluated on EvTTC without fine-tuning, demonstrates superior robustness. It achieves the best accuracy on both CCRs2-medium (8.31\%) and CCRs2-high (10.56\%), consequently securing the best average RTE (10.60\%). This strongly suggests our method possesses superior generalization capabilities beyond simple car-following models. Critically, our method also achieves the fastest average runtime (12.67~ms), which is nearly twice as fast as the next-best competitor, STRTTC (25.00~ms).}

\begin{figure}[t]
    \centering
    \includegraphics[width=0.47\textwidth]{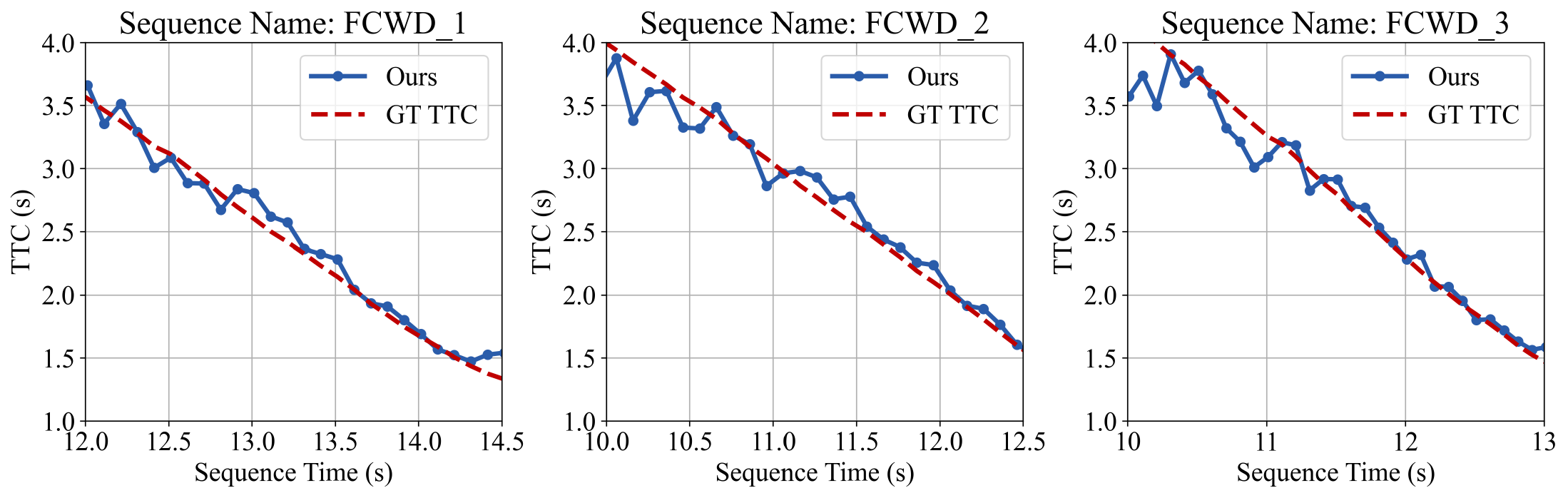}
    \caption{Comparison of predicted TTC with ground truth on the FCWD benchmark~\cite{li2024strttc} without model fine-tuning.}
    \label{fig:generalization_experiment}
\end{figure}

\rebuttal{
This demonstrated generalization stems directly from the representation learning approach fostered by our \emph{eAP} dataset, which was designed to facilitate such dedicated studies. The generality of this learned representation is further validated by a cross-dataset experiment on the FCWD benchmark~\cite{li2024strttc}. As shown in Fig.~\ref{fig:generalization_experiment} and Tab.~\ref{tab:method_benchmark_fcwd}, our proposed approach also outperforms the baseline on FCWD. Collectively, these results (EvTTC and FCWD) confirm that our event-enhanced TTC estimation model not only surpasses previous approaches using either event or image data but also achieves a compelling balance of accuracy, generalization, and real-time performance, proving its suitability for robust, real-world robotic applications.
}

\vspace{1ex}

\begin{table}[t] %
\centering
\setlength{\tabcolsep}{3pt} 
\resizebox{0.45\textwidth}{!}{%
\normalfont
\begin{tabular}{lrrrrr} 
\toprule
& 
& \textbf{\emph{FCWD1$_{\text{RTE}}$}}$\downarrow$ 
& \textbf{\emph{FCWD2$_{\text{RTE}}$}}$\downarrow$ 
& \textbf{\emph{FCWD3$_{\text{RTE}}$}}$\downarrow$ 
& \textbf{\emph{eAP$_{\text{MiD}}$}}$\downarrow$ \\

\midrule
\multirow{2}{*}{\textbf{\makecell[c]{Event-based}}} 
& ETTCM$_{\text{6-dof}}$~\cite{nunes2023time} & 15.5 & 18.4 & 19.0 & 265.4 \\
& STRTTC~\cite{li2024strttc} & 9.8 & 11.5 & 14.0 & 408.7 \\
\midrule
\textbf{Frame-based} 
& DeepScale~\cite{TSTTC} & 25.4 & 19.6 & 21.7 & 81.9 \\

\midrule

\multirow{3}{*}{\textbf{Ours}} 
& Visual-only baseline & 48.5 & 47.6 & 46.8 & 160.6 \\
& Visual-only + LHR & 13.3 & 16.7 & 16.6 & 68.3 \\
& Ours (Full) & \textbf{5.2} & \textbf{6.1} & \textbf{5.4} & \textbf{45.0} \\
\bottomrule
\end{tabular}
}
\caption{Quantitative evaluation of object TTC estimation performance \rebuttal{on the FCWD benchmark~\cite{li2024strttc} and the~\emph{eAP} dataset.}}
\label{tab:method_benchmark_fcwd}
\vspace{-7ex}
\end{table}

\vspace{1ex}

\begin{table*}[t]
\centering
\resizebox{\linewidth}{!}{%
\normalfont
\begin{tabular}{lllllllrrrrr}
\toprule
\multicolumn{1}{c}{\multirow{2}{*}{Experimental Setting}} &
  \multicolumn{2}{c}{Input Modality} &
  \multicolumn{2}{c}{Approach} &  
  \multicolumn{2}{c}{Fusion Strategy} &
  \multirow{2}{*}{MiD$_{\text{c}}\downarrow$} &
  \multirow{2}{*}{MiD$_{\text{s}}\downarrow$} &
  \multirow{2}{*}{MiD$_{\text{l}}\downarrow$} &
  \multirow{2}{*}{MiD$_{\text{n}}\downarrow$} &
  \multirow{2}{*}{MiD$_{\text{overall}}\downarrow$} \\ 
\cmidrule(l{1mm}r{1mm}){2-3}
\cmidrule(l{1mm}r{1mm}){4-5}
\cmidrule(l{1mm}r{1mm}){6-7}
\multicolumn{1}{c}{} &
  \multicolumn{1}{c}{RGB} &
  \multicolumn{1}{c}{Event} &
  \multicolumn{1}{l}{LHR} &
  \multicolumn{1}{l}{FS} &
  \multicolumn{1}{l}{Early} &
  \multicolumn{1}{l}{Late} &
   &
   &
   &
   &
   \\ \midrule

Visual-only baseline &
  \multicolumn{1}{l}{\cmark} &
  \xmark &
  \multicolumn{1}{l}{\xmark} &
  \xmark &
  \multicolumn{1}{l}{-} & 
   - &
  222.9 &
  105.7 &
  93.9 & 
  80.4 &
  160.6
  \\ 
Event-only baseline &
  \multicolumn{1}{l}{\xmark} &
  \cmark &
  \multicolumn{1}{l}{\xmark} &
  \xmark &
 \multicolumn{1}{l}{-} & 
   - &
   93.9 & 
   65.2 &
   89.6 & 
   42.4 &
   79.7
   \\ 
Multi-modal baseline &
  \multicolumn{1}{l}{\cmark} &
  \cmark &
  \multicolumn{1}{l}{\xmark} &
  \xmark &
  \multicolumn{1}{l}{\cmark} & 
   \xmark &
   164.9&
   97.6&
   117.5&
   66.1&
   130.1
   \\ 
Visual-only + LHR &
  \multicolumn{1}{l}{\cmark} &
  \xmark &
  \multicolumn{1}{l}{\cmark} &
  \xmark &
  \multicolumn{1}{l}{-} & 
   - &
   95.6 &
   47.4 &
   30.5 &
   32.4 &
   68.3
   \\ 
Event-only + LHR &
  \multicolumn{1}{l}{\xmark} &
  \cmark &
  \multicolumn{1}{l}{\cmark} &
  \xmark &
  \multicolumn{1}{l}{-} & 
   - &
   90.5 &
   44.8 &
   43.0 &
   32.2 &
   66.2
   \\ 
Multi-modal + LHR &
  \multicolumn{1}{l}{\cmark} &
  \cmark &
  \multicolumn{1}{l}{\cmark} &
  \xmark &
  \multicolumn{1}{l}{\cmark} & 
  \xmark &
   94.9 &
   49.6 &
   37.6 &
   36.5 &
   69.7
  \\ 
Multi-modal + LHR &
  \multicolumn{1}{l}{\cmark} &
  \cmark &
  \multicolumn{1}{l}{\cmark} &
  \xmark &
  \multicolumn{1}{l}{\xmark} & 
  \cmark &
   70.4 & 
   35.6 & 
   34.8 & 
   36.7 & 
   53.0
  \\  \midrule
Ours (Full) &
  \multicolumn{1}{l}{\cmark} &
  \cmark &
  \multicolumn{1}{l}{\cmark} &
  \cmark &
  \multicolumn{1}{l}{\xmark} & 
  \cmark &
   \textbf{53.1} & 
   \textbf{37.6} &
   40.6 &
   \textbf{31.3} &
   \textbf{45.0}
   
   \\ \bottomrule
\end{tabular}
}
\caption{Motion-in-depth error (MiD) on the testing split for different ranges of TTC ground truth \rebuttal{on the \emph{eAP} testing set}. In each experiment, \cmark~indicates a proposed technique is used and \xmark~indicates otherwise. LHR: height-ratio-aware representation learning. FS: foreground supervision.}
\label{tab:ablation}
\end{table*}

\subsubsection{Ablation study}

We also conduct extensive ablation studies \rebuttal{on the \emph{eAP} dataset} to assess the contribution of each technique. 
The results, summarized in Tab.~\ref{tab:ablation}, are obtained under consistent experimental configurations across all studies.

\noindent\textbf{TTC regression from frame vs. event inputs.} Comparing the baseline models with different input modalities, the event-based input achieves a significantly lower MiD of 79.7 compared to 160.6 for frame-based input. 
We attribute this improvement to the rich temporal information encoded in event data, which captures changes in object silhouettes more effectively.

\noindent\textbf{Effectiveness of learning geometry-aware representations.} The baseline model, which directly regresses TTC from RGB and event inputs, achieves a MiD of 160.6 and 79.7, respectively, highlighting the inherent difficulty of the learning task. 
In contrast, incorporating geometry-aware representations that account for changes in object heights yields significant improvements. 
For frame-only models, learning height ratios reduces MiD from 160.6 to 68.3, demonstrating that visual object heights simplify the nonlinear mapping from RGB pixels to TTC. 
Similar improvements are observed for models using event data or multi-modal inputs. By guiding the model to learn geometry-aware features during training, it generalizes more effectively to unseen scenarios.

\noindent\textbf{Effectiveness of different modalities for different scenes.} 
Comparing height-ratio networks with different inputs, we observe that event-based networks out-perform frame-only models in scenes with small TTC ground truths. 
This is attributed to the high relative speed between the object and the ego vehicle, which generates a large number of informative events. 
On the other hand, in scenes with large TTC ground truths, the frame-only model performs better due to the limited number of events in low relative speed scenarios.

\noindent\textbf{Early-fusion vs. late-fusion.} 
We further train multi-modal variants using two fusion strategies: early-fusion, which concatenates frames and event volumes for joint feature extraction (Fig.~\ref{fig:model_architecture}(a--b)); and late-fusion, which independently extracts features from both modalities before concatenating high-level representations (Fig.~\ref{fig:model_architecture}(c--d)). 
As shown in Tab.~\ref{tab:ablation}, the late-fusion strategy out-performs the early-fusion one for both baseline and height-ratio networks, indicating the modality gap between frames and event data. 
The multi-modal model achieves a MiD of 53.1 for \emph{crucial} scenarios and an overall MiD of 45.0, surpassing single-modal height-ratio networks and demonstrating its ability to leverage complementary strengths of frame and event data for optimal TTC estimation.

\begin{figure}[t]
    \centering
    \includegraphics[width=0.47\textwidth]{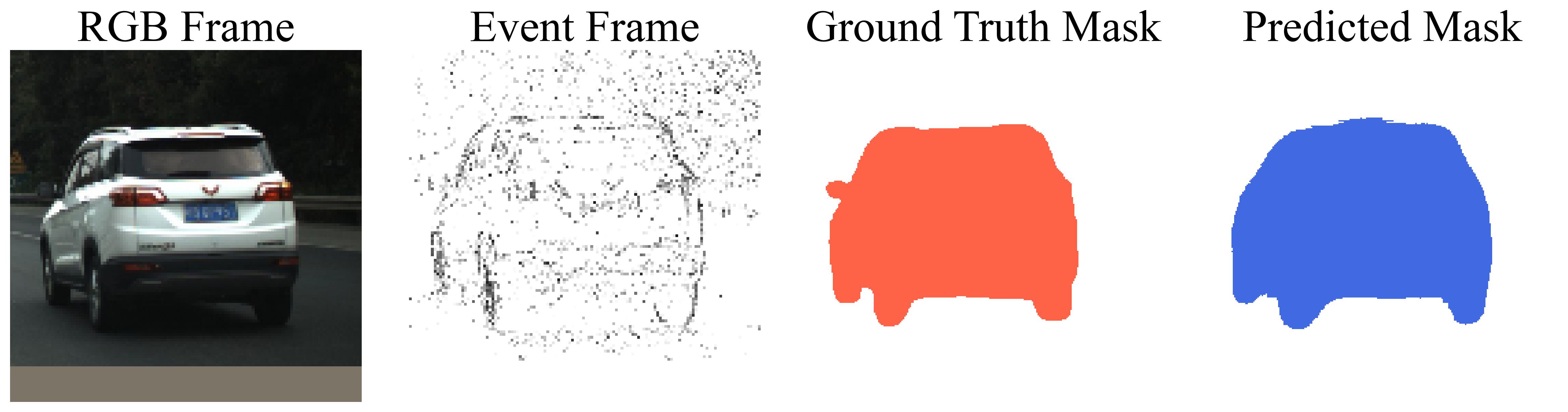}
    \caption{The predicted mask of our student model (blue) and that of \rebuttal{SAM~\cite{kirillov2023segment} (red)} used as ground truth for supervision.}
    \label{fig:pred_mask}
    \vspace{-1ex}
\end{figure}

\noindent\textbf{Benefits of learning foreground-aware representations.}
Adding the decoder and object foreground supervision lowers the MiD from 53.0 to 45.0, showing that enhancing the model's ability to discern fine-grained object boundaries improves TTC estimation. 
\rebuttal{As illustrated in Fig.~\ref{fig:pred_mask}, the predicted masks align closely with references generated by the teacher model, highlighting the network's capability to reconstruct vehicle contours from event feature maps accurately.} This enables precise TTC estimation in dynamic scenarios.

\subsubsection{Deployment on Edge Device}
Without optimization of network architecture or quantization, our event-based object TTC estimation network runs at 200~FPS on an NVIDIA Orin NX GPU \rebuttal{with 16~GB VRAM, taking event and image data from vehicle Regions of Interest (ROIs)—identified by 2D bounding box detections—which are then resized to 128x128 as input, and} using the TensorRT \rebuttal{library}\footnote{https://developer.nvidia.com/tensorrt} for deployment. This model is the first deep network that achieves state-of-the-art performance for event-based object TTC estimation trained with a large-scale dataset. It demonstrates the possibilities of deep representation learning from an increasing amount of collected event data.

\rebuttal{
\subsection{Runtime analysis}}

\vspace{1ex}
\begin{table}[t]
\centering
\setlength{\tabcolsep}{3pt}
\resizebox{0.48\textwidth}{!}{
\begin{tabular}{l l l r r}
    \toprule
    Model & Device & Module & Avg. (ms) & Total (ms) \\
    \midrule
    
    \multirow{5}{*}{Garl-TTC} & 
    \multirow{3}{*}{\makecell[l]{NVIDIA A100 \\ (PyTorch)}} & 
    RGB Encoder & 7.11 & \multirow{3}{*}{21.05} \\
     & & Event Encoder & 7.08 & \\
     & & Height Head & 0.15 & \\
    \cmidrule{2-5} 
     & \makecell[l]{NVIDIA Orin NX16G \\ (ONNX, MaxN)} & 
    All Modules & 4.55 & 4.55 \\
    
    \midrule
    
    \multirow{6}{*}{3D Det.} & 
    \multirow{6}{*}{\makecell[l]{NVIDIA A100 \\ (PyTorch)}} & 
    RGB Encoder & 27.48 & \multirow{6}{*}{125.07} \\
     & & Event Encoder & 23.35 & \\
     & & Fusion Module & 0.78 & \\
     & & View Transformer & 7.96 & \\
     & & BEV Encoder & 34.26 & \\
     & & Detection Head & 1.68 & \\
    
    \bottomrule
\end{tabular}
}

\captionof{table}{\rebuttal{Runtime analysis of different model components (in ms).}}
\label{tab:runtime_analysis}
\vspace{-1ex}
\end{table}

\jhang{We analyze the computational performance of our proposed models on two distinct hardware platforms: a high-end NVIDIA A100-40G GPU for baseline profiling using PyTorch, and an embedded NVIDIA Orin NX 16~GB module for real-world deployment evaluation. To ensure maximum performance on the edge device, the Orin NX was configured to its maximum performance (MAXN) mode, allowing for its peak 25~W power draw and highest clock frequencies.}

\jhang{As detailed in Tab.~\ref{tab:runtime_analysis}, the Garl-TTC model demonstrates high efficiency. In its native PyTorch format on the A100-40G, the model achieves a total inference time of 21.05~ms. A module-level breakdown reveals that the primary computational load stems from the dual-stream backbones: the \textit{RGB Encoder} (7.11~ms) and the \textit{Event Encoder} (7.08~ms). The final \textit{Height Regressor} has a negligible inference time of 0.15~ms.
Crucially, when optimized for deployment via ONNX conversion and executed on the Orin NX (MAXN Mode), the Garl-TTC model achieves a total inference time of just 4.55~ms. This runtime translates to approximately 220~FPS, confirming the model is highly efficient and well-suited for real-time autonomous systems where low latency is critical.}

\jhang{In contrast, the full 3D Detection model is significantly more computationally intensive, reflecting its architectural complexity. On the A100-40G, this model requires 125.07~ms for a single forward pass. The analysis identifies the \textit{BEV Encoder} (34.26~ms) as the most demanding component, followed by the \textit{RGB Encoder} (27.48~ms) and the \textit{Event Encoder} (23.35~ms). The \textit{View Transformer} (LSS) module also introduces a notable 7.96~ms of latency. Given this substantial computational load, deploying the 3D detection model on an edge device would require further optimization techniques, such as quantization or model pruning.}

\rebuttal{
\subsection{Limitations}}
While our proposed methods yield promising results on the \emph{eAP} dataset, several limitations present clear avenues for future research. For instance, the Garl-TTC framework exhibits sensitivity to ego-motion, particularly high rotation. Its geometric formulation (Eq. 11), which relies on the ratio of visual object heights, inherently assumes that changes in visual scale are primarily due to translational looming. Consequently, rotational ego-motion can introduce pixel displacements that are misinterpreted as translational motion, affecting TTC accuracy. A clear direction to enhance robustness is to integrate ego-motion compensation, potentially by fusing IMU data to decouple these motion components. Regarding the event-enhanced 3D vehicle detection task, challenges also remain. As discussed in Sec. VII-B.1, event sparsity in low relative-speed scenarios limits the event modality's contribution. Furthermore, our current feature concatenation approach may not fully exploit the high temporal resolution of event data. Therefore, future work could explore more sophisticated model architectures, such as deeper or asynchronous fusion mechanisms, to better leverage the high-speed nature of event data.
\final{Furthermore, while we focus on open-loop perception, real-world vehicles inherently face action latencies and mechanical constraints~\cite{gehrig2024low}. Our study does not attempt to identify the minimum sensing rate required for this task; rather, it shows that low-latency perception is practically relevant and motivates future closed-loop studies on when higher-frequency sensing yields measurable benefits.}
We hope that our \emph{eAP} dataset will serve as a foundational benchmark for the event-enhanced visual perception community, supporting and catalyzing the development of more advanced algorithms for both TTC estimation and event-enhanced 3D perception.

\vspace{4ex}
\section{Conclusion}
\label{sec:conclusion}

We present a new large-scale dataset, the \emph{eAP} dataset, which features event cameras and encompasses a diverse range of autonomous driving scenes. \emph{eAP} enables an in-depth study of deep representation learning for event-enhanced visual autonomous perception. As one example, \emph{eAP} leads to the first discovery of how event cameras can benefit a modern BEV 3D vehicle detection approach. As another example, \emph{eAP} facilitates an in-depth study of geometry-aware representation learning for event-based object TTC estimation, enabling a real-time deep model with state-of-the-art accuracy. This work lays the foundation for future studies that can explore advanced multi-modal fusion strategies, long-term temporal representation learning, and improved edge-device deployment, thereby advancing real-world autonomous perception using event cameras. \rebuttal{Future efforts also include installing the data acquisition system onto other robotic platforms (such as quadrotors or quadrupeds), so that the scope of this benchmark can cover more scenarios beyond the autonomous driving environment.}

\section*{Acknowledgment}
\label{sec:ack}
The authors would like to thank Mr. Yuhao Liu for his guidance on camera hardware design, Mr. Zhaoxia Ruan for his guidance on camera parameter tuning, and Mr. Haoyuan Pei for his assistance with data annotation.
The authors also thank Mr. Huachun Wang, Mr. Ye Wang, and Mr. Weiliang Chen for their assistance with data collection.
The authors further thank Dr. Yi Yu for proofreading.
This work was supported in part by the National Natural Science Foundation of China under Grant 62427813, and in part by the National Key Research and Development Project of China under Grant 2023YFB4706600.

\bibliographystyle{IEEEtran} %
\bibliography{myBib}

\raggedbottom
\begin{IEEEbiography}[{\includegraphics[width=1in,height=1.25in,clip,keepaspectratio]{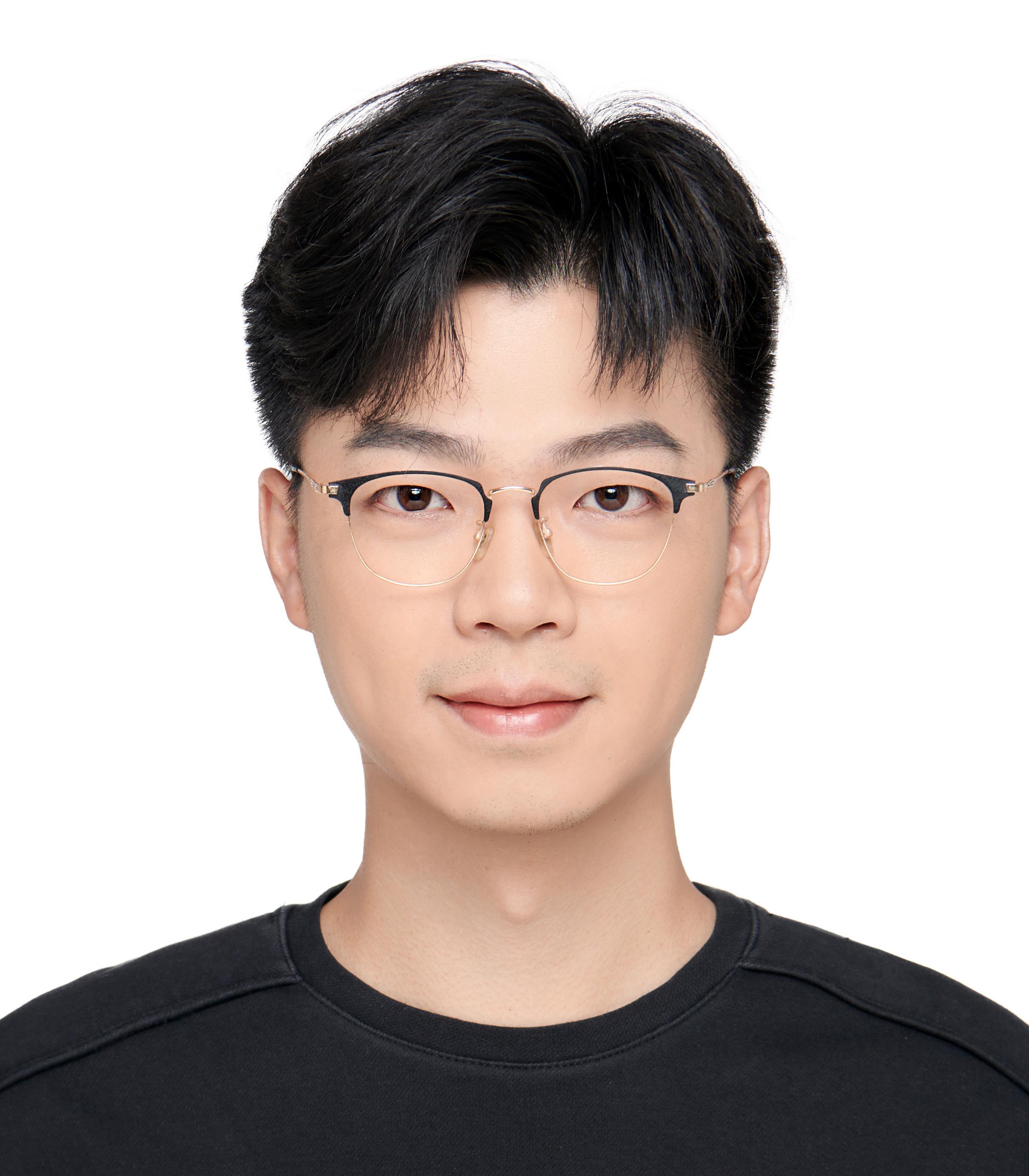}}]{Jinghang Li}
    is currently pursuing the Ph.D. degree at the College of Artificial Intelligence and Robotics, Hunan University, Changsha, China, under the supervision of Prof. Yi Zhou. He received the M.Eng. degree from the Beijing Institute of Technology, Beijing, China. His research interests include 3D perception, multimodal foundation models for autonomous driving, event-based vision, data-centric learning, and geometry-aware modeling.
\end{IEEEbiography}

\begin{IEEEbiography}[{\includegraphics[width=1in,height=1.25in,clip,keepaspectratio]{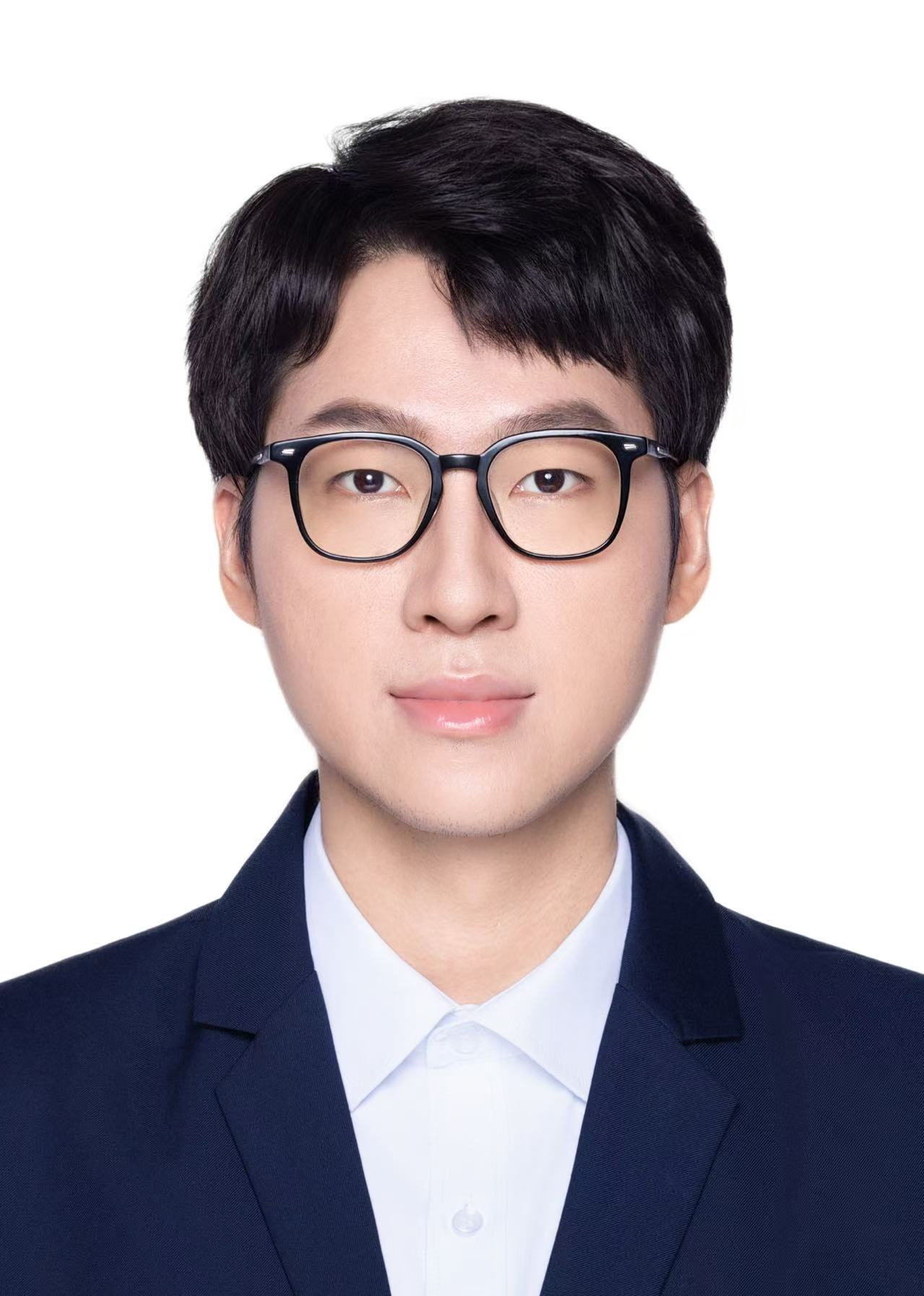}}]{Shichao Li}
    received the B.E. degree in information engineering from Zhejiang University, Zhejiang, China, in 2017, and the Ph.D. degree in computer science and engineering from The Hong Kong University of Science and Technology, Hong Kong, in 2022.
    He is currently a Senior Research Engineer at ByteDance. He previously served as a Senior Computer Vision Engineer and project lead at DJI. His research interests include computer vision, machine learning, and representation learning as well as their applications in 3D perception, multi-modal understanding, generation and simulation.
\end{IEEEbiography}

\begin{IEEEbiography}[{\includegraphics[width=1in,height=1.25in,clip,keepaspectratio]{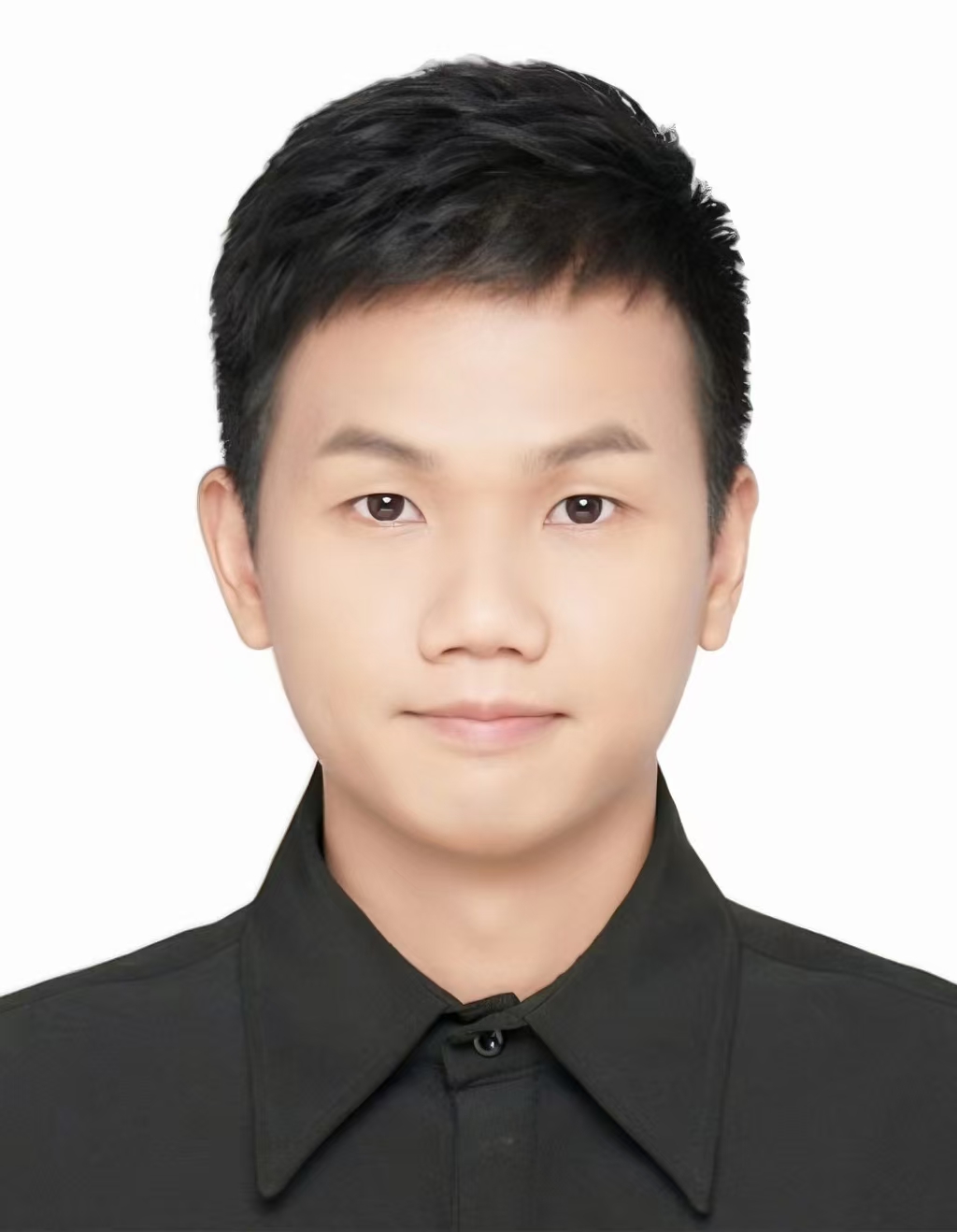}}]{Qing Lian}
    received the Ph.D. degree in computer science from the Hong Kong University of Science and Technology, Hong Kong, in 2023. He is currently a Researcher with the Computer Vision and Robotics Group, IDEA Research. His research interests include vision-language-action models, reinforcement learning, 3D vision, and their applications in self-driving and embodied AI.
\end{IEEEbiography}

\begin{IEEEbiography}[{\includegraphics[width=1in,height=1.25in,clip,keepaspectratio]{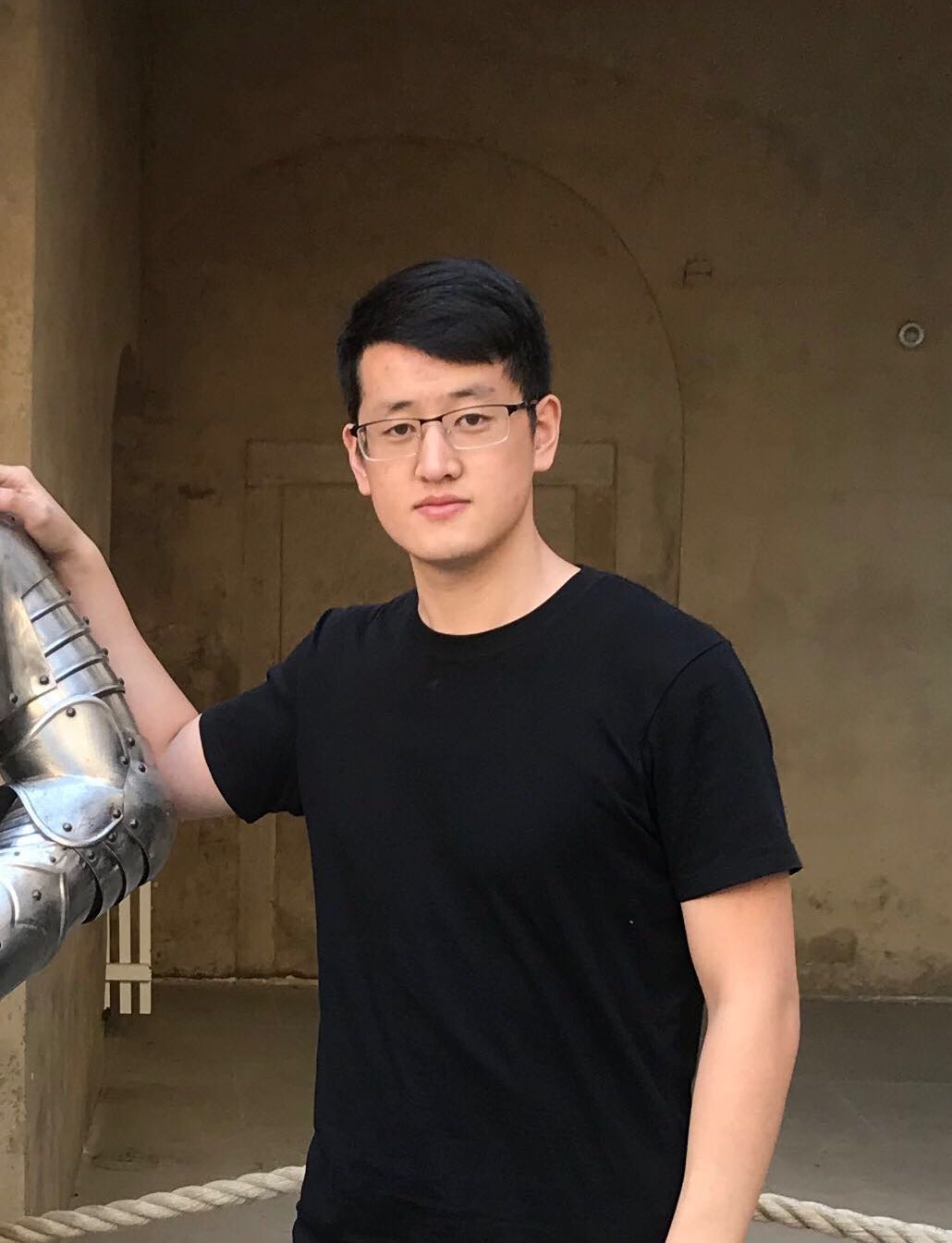}}]{Peiliang Li}
    received the Ph.D. degree from the Hong Kong University of Science and Technology (HKUST), supervised by Prof. Shaojie Shen. He currently leads the end-to-end self-driving algorithms and the next-generation algorithm team at ZYT.
\end{IEEEbiography}

\begin{IEEEbiography}[{\includegraphics[width=1in,height=1.25in,clip,keepaspectratio]{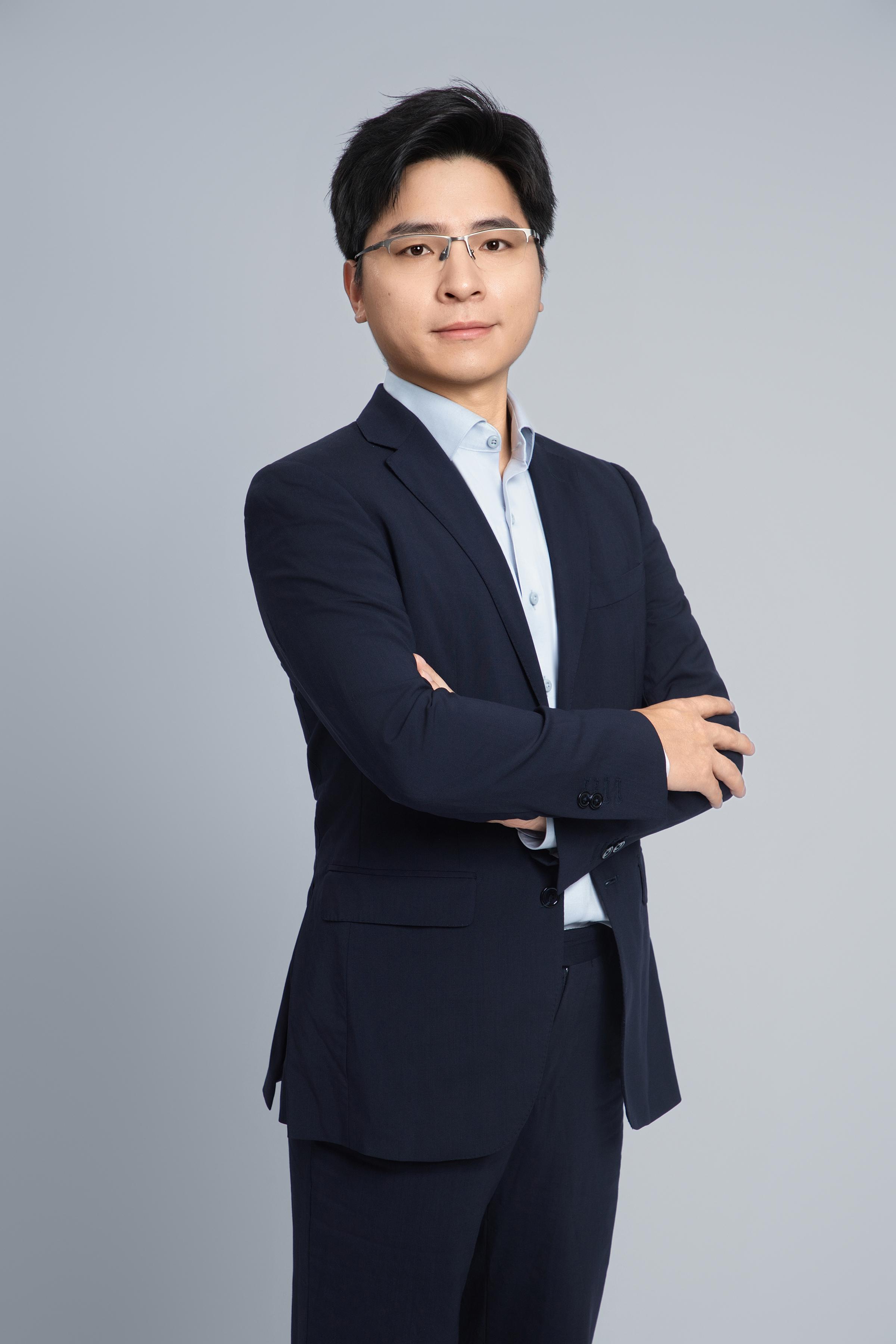}}]{Xiaozhi Chen}
    received the Ph.D. degree in electronic engineering from Tsinghua University, Beijing, China, in 2017. He is currently the director of AI research at ZYT. His research interests include foundation models, autonomous driving and robotics.
\end{IEEEbiography}

\begin{IEEEbiography}[{\includegraphics[width=1in,height=1.25in,clip,keepaspectratio]{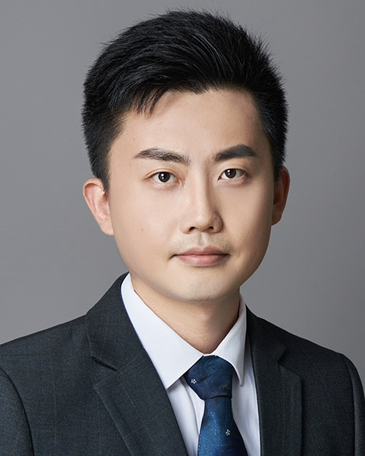}}]{Yi Zhou} (Member, IEEE) is a Professor at Hunan University, where he directed the Neuromorphic Automation and Intelligence Lab (NAIL).
He obtained his Ph.D. degree in engineering and computer science from the Australian National University, Canberra, Australia in 2018.
He was a visiting scholar at ETH Zurich (2017-2018) and was awarded the NCCR Fellowship Award by the Swiss National Science Foundation for his research on neuromorphic event-based 3D vision. 
From 2019 to 2021, he was a postdoc research fellow at the HKUST-DJI Innovation Joint Lab, where he proposed the world's first open-source event-based stereo visual odometry (ESVO) system.
His research interests include visual odometry / simultaneous localization and mapping (SLAM), geometry problems in computer vision, and dynamic vision sensors. 
His team won the championship in the SLAM Challenge at the CVPR 2025 Workshop on Event-Based Vision in Nashville.
He is currently serving as an associate editor for IEEE RA-L and IEEE T-RO.
\end{IEEEbiography}

\vfill

\end{document}